\documentclass{article} 
\usepackage{2027_conference,times}

\usepackage{amsmath,amsfonts,bm}

\def\eqref#1{equation~\ref{#1}}

\def\1{\bm{1}}

\DeclareMathAlphabet{\mathsfit}{\encodingdefault}{\sfdefault}{m}{sl}
\SetMathAlphabet{\mathsfit}{bold}{\encodingdefault}{\sfdefault}{bx}{n}

\usepackage{hyperref}
\usepackage{url}
\newcommand{\MechTopOneCoverage}{96.97\%}

\newcommand{\MechIntervention}{43.07\%}
\newcommand{\MechRescue}{28.77}
\newcommand{\MechRegression}{1.60}

\newcommand{\MechDeltaBlockLogP}{7.63}

\newcommand{\MechAdjacentIncidence}{2.22\%}
\newcommand{\MechAdjacentBreakerShare}{10.02\%}
\newcommand{\MechNonadjacentIncidence}{3.94\%}
\newcommand{\MechNonadjacentBreakerShare}{4.27\%}

\newcommand{\MechAdjacentHeadRescue}{57.32\%}

\newcommand{\MechNonadjacentHeadRescue}{34.80\%}
\newcommand{\MechBlockPositive}{33.6\%}
\newcommand{\MechBlockTied}{52.3\%}
\newcommand{\MechBlockNegative}{14.2\%}

\usepackage{booktabs}
\usepackage{makecell}
\usepackage{adjustbox}
\usepackage{graphicx}
\usepackage{algorithm}
\usepackage[noend]{algpseudocode}
\usepackage{xcolor}
\usepackage{amsmath}
\usepackage{amssymb}
\usepackage[table]{xcolor} 
\usepackage{array} 
\usepackage{makecell} 
\usepackage{adjustbox}
\usepackage{tabularx}

\title{From Position Risks to Block Survival: Faster Generation for Diffusion Language Models}

\author{
\textbf{Siwei Chen}$^{*,1}$ \quad
\textbf{Yuxiang Wan}$^{*,1}$ \quad
\textbf{Yifan Yu}$^{1}$ \quad
\textbf{Fan Lai}$^{1}$
\\[5pt]
{\normalsize $^{1}$University of Illinois Urbana-Champaign}
\quad
{\small $^{*}$Equal contribution}
}

\usepackage{xspace}
\def\name{BRISK-DLM\xspace}

\usepackage[inline]{enumitem}
\newenvironment{denseitemize}{
\begin{itemize}[topsep=2pt, partopsep=0pt, leftmargin=1.5em]
  \setlength{\itemsep}{2pt}
  \setlength{\parskip}{0pt}
  \setlength{\parsep}{0pt}
}{\end{itemize}}

\begin{document}

\maketitle

\begin{abstract}
Diffusion language models (DLMs) can accelerate generation by predicting multiple tokens in parallel, but there is a mismatch between how these tokens are predicted and how they ultimately contribute to generation. Parallel predictions can hardly condition on the tokens selected earlier within the same block, even though their validity depends on this realized prefix. Under the popular proposal--verification decoding, this mismatch makes errors highly asymmetric: an early rejection prevents all subsequent proposals from contributing decoding progress. We introduce BRISK-DLM, a framework that addresses both mismatches by optimizing proposal learning and selection for verified progress.  BRISK-DLM trains on self-generated sequences, using risk–reward weighting to dynamically prioritize positions by their impact on verified progress and decoding cost. During inference, a lightweight prefix-conditioned corrector reranks existing candidates using previously selected tokens and preferences distilled from the model’s own verifier. The corrector reuses the backbone's parallel representations and requires no additional backbone evaluation, while fused execution keeps its overhead small. 
BRISK-DLM improves end-to-end throughput by up to 37.4\% while preserving task quality, establishing a new quality--throughput frontier for DLM generation.

\end{abstract}

\section{Introduction}

Autoregressive language models generate one token at a time, making sequential dependence a fundamental bottleneck for inference. Diffusion language models (DLMs) offer a different computational structure: multiple future token positions can be predicted within a single model evaluation. Although this capability is realized differently across discrete diffusion, masked prediction, and block-based formulations, recent DLMs have scaled to general-purpose language modeling and reasoning while substantially narrowing the quality gap with autoregressive models
\citep{austin2021d3pm,sahoo2024mdlm,lou2024sedd,nie2025llada,arriola2025blockdiffusion}.
This raises an appealing possibility: language generation need not be tied to one-token-at-a-time computation.

Yet predicting tokens in parallel is not the same as making progress in parallel. In DLMs, future positions are predicted simultaneously, before the token choices at earlier positions within the same block are known, even though the appropriate prediction at a later position depends on this realized prefix. Under the increasingly popular proposal--verification decoding\citep{chen2023speculative}, this mismatch becomes especially costly: proposals are verified from left to right, so an early rejection prevents all subsequent proposals from contributing immediate decoding progress. Therefore, parallel generation faces two fundamental consequences of prefix dependence: \emph{which positions are worth improving} depends on how much downstream progress they offer, while \emph{which tokens should be selected} depends on the prefix realized before them.
This perspective exposes a gap in existing approaches to parallel DLM generation. Prior work improves model capability\citep{gong2024scaling}, block-parallel generation\citep{fu2025efficientdlm}, specialized decoding\citep{arriola2025blockdiffusion}, token editing\citep{bie2026llada21}, caching\citep{wu2025fastdllm}, and acceptance-aware training
\citep{wu2026dpace}.
However, proposal learning and within-block token selection are typically optimized separately, even though both determine how much parallel computation becomes usable output. 

Instead, we argue that DLM efficiency should be organized around a common principle: \emph{verified progress per model evaluation}. Achieving this requires improving both the proposal distribution itself and how its candidates are selected. This raises two integral questions. During learning, how should the model allocate its capacity across proposal positions? A position is more valuable to improve when it is likely to fail and when its survival unlocks substantial downstream progress. During decoding, which candidate should be selected? A token that is locally plausible under parallel prediction may become inappropriate once earlier choices within the block are realized. 

\begin{figure}[t]
    \centering
    \begin{minipage}[t]{0.48\linewidth}
        \centering
        \vspace{0pt}
        \includegraphics[width=\linewidth]
        {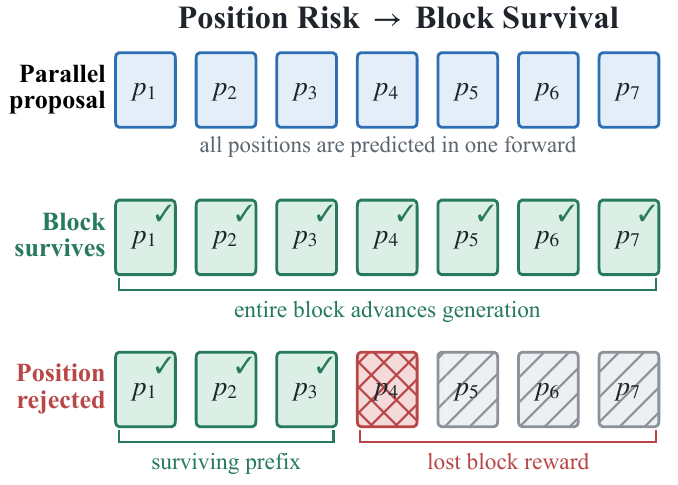}
    \end{minipage}
    \hfill
    \begin{minipage}[t]{0.49\linewidth}
        \centering
        \vspace{0pt}
        \includegraphics[width=\linewidth]
        {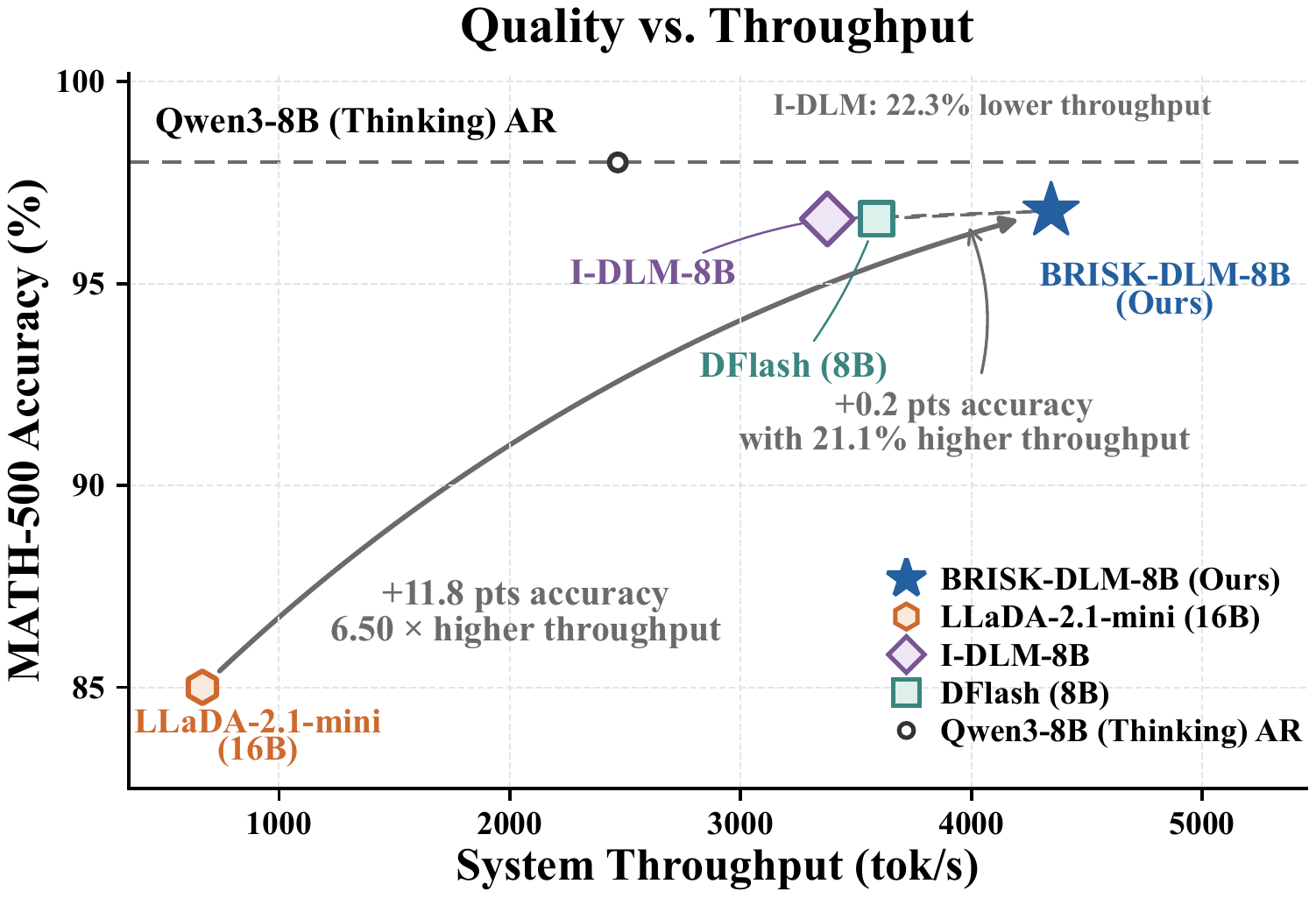}
    \end{minipage}
    \vspace{-.2cm}
    \caption{\textbf{From position risks to the quality--throughput frontier.}
    \textbf{Left:} Parallel proposals advance generation only through the consecutively accepted prefix.
    \textbf{Right:} BRISK-DLM shifts the quality--throughput frontier toward higher throughput while preserving quality.
    }
    \label{fig:intro_overview}
    \vspace{-12pt}
\end{figure}

Based on this principle, we introduce \textbf{BRISK-DLM}, a framework that aligns both proposal learning and candidate selection with verified decoding progress. During training, BRISK-DLM learns from correctness-filtered trajectories generated under the deployed decoding policy and assigns learning signal according to each proposal position's marginal effect on decoding efficiency. This risk--reward aware objective concentrates model capacity on the errors that currently limit block progress, while shifting emphasis as the acceptance profile evolves. During inference, BRISK-DLM complements stronger proposals with a lightweight prefix-conditioned corrector that reranks existing candidates using the tokens already selected within the block. The corrector is trained to reproduce preferences from the model's own verifier on states encountered by the deployed proposal policy. Importantly, the DLM backbone remains fully parallel: the corrector reuses its representations and candidate set, requires no additional backbone evaluation, and is integrated through fused execution so that improved prefix survival translates into end-to-end speedup. Figure~\ref{fig:intro_overview} connects this position-to-block mechanism to the ultimate objective of improving throughput without sacrificing quality.

Overall, this paper makes the following contributions:

\begin{denseitemize}
\item We identify verified progress per model evaluation as a unifying objective for efficient parallel DLM generation. This view connects two challenges that are typically treated separately: how learning capacity should be allocated across proposal positions, and how parallel candidates should be selected once the within-block prefix is realized.

\item We introduce BRISK-DLM to address both challenges. A risk--reward objective dynamically prioritizes proposal positions according to their marginal effect on verified progress, while a lightweight prefix-conditioned corrector distills verifier preferences on policy-reached states to assemble more compatible proposal blocks without additional backbone evaluations.

\item We evaluate BRISK-DLM across different model architectures, scales, and reasoning tasks under diverse serving configurations.
BRISK-DLM increases verified tokens per forward by up to 37.4\% over existing advances, including DFlash~\citep{chen2026dflash}, while preserving task quality.
These results unlock a new quality--throughput frontier for DLM inference.

\end{denseitemize}

\section{Related Work and Motivation}

\subsection{Background and Related Work}
\label{sec:idlm_background}

At each step, the DLM predicts token distributions for multiple unresolved positions in parallel, and a decoding schedule determines which predictions are committed or revised before the next step. Masked DLMs apply this process across an entire response, while block-based variants maintain a clean prefix and resolve one block (e.g., 16 tokens) at a time \citep{austin2021d3pm,sahoo2024mdlm,nie2025llada,arriola2025blockdiffusion}. While  converting one model evaluation into progress on multiple tokens, parallel prediction also reduces causal information: later positions are computed before the actual tokens selected at earlier positions are known. This creates an opportunity for lightweight prefix-conditioned refinement at inference, without sacrificing parallel backbone computation.
 
\paragraph{Verified parallel generation.}
A growing class of DLM uses proposal--verification to convert parallel predictions into output, including DFlash~\citep{chen2026dflash,wu2026dpace}, self-speculative block-diffusion method such as SSD, FreeDave, and S2D2~\citep{gao2025selfspeculativedecodingdiffusion,wu2025free,han2026s2d2}, and I-DLM with introspective strided decoding (ISD)~\citep{yu2026idlm}. Although implementations differ, they share a common structure: multiple future tokens are proposed in parallel, while a verifier determines how much can be committed. An early rejection prevents later proposals from contributing immediate progress, making positions unequal in their effect on decoding efficiency. Meanwhile, verification evaluates each candidate under a realized prefix unavailable when the parallel proposal was produced. We provide a detailed related work in Appendix~\ref{related}.

\begin{figure}[t]
    \centering
    \begin{minipage}[t]{0.37\linewidth}
        \centering
        \vspace{0pt}
        \includegraphics[width=\linewidth]
        {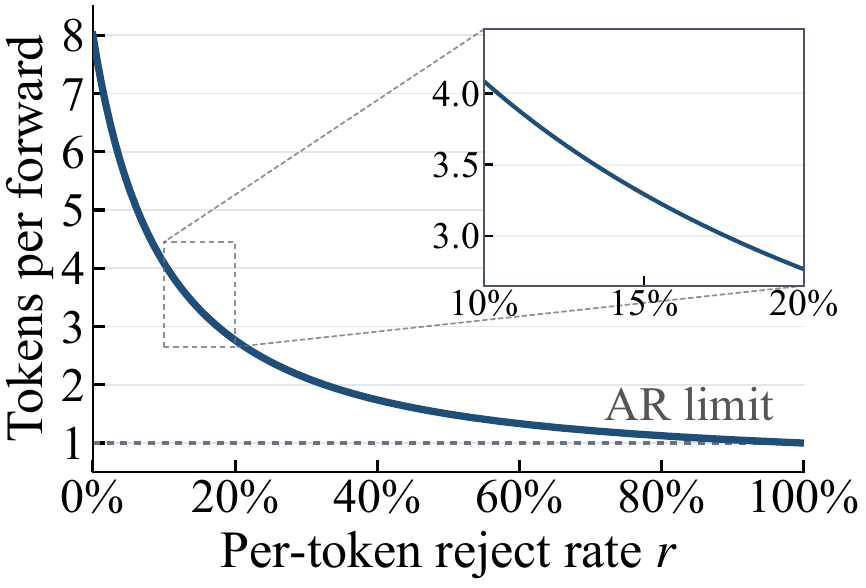}
        \vspace{-.5cm}
        \caption{
        \textbf{Rejection limits effective parallelism.} Theoretical TPF at stride $N=8$.
        }
        \label{fig:tpf_rejection}
    \end{minipage}
    \hfill
    \begin{minipage}[t]{0.62\linewidth}
        \centering
        \vspace{0pt}
        \includegraphics[width=\linewidth]
        {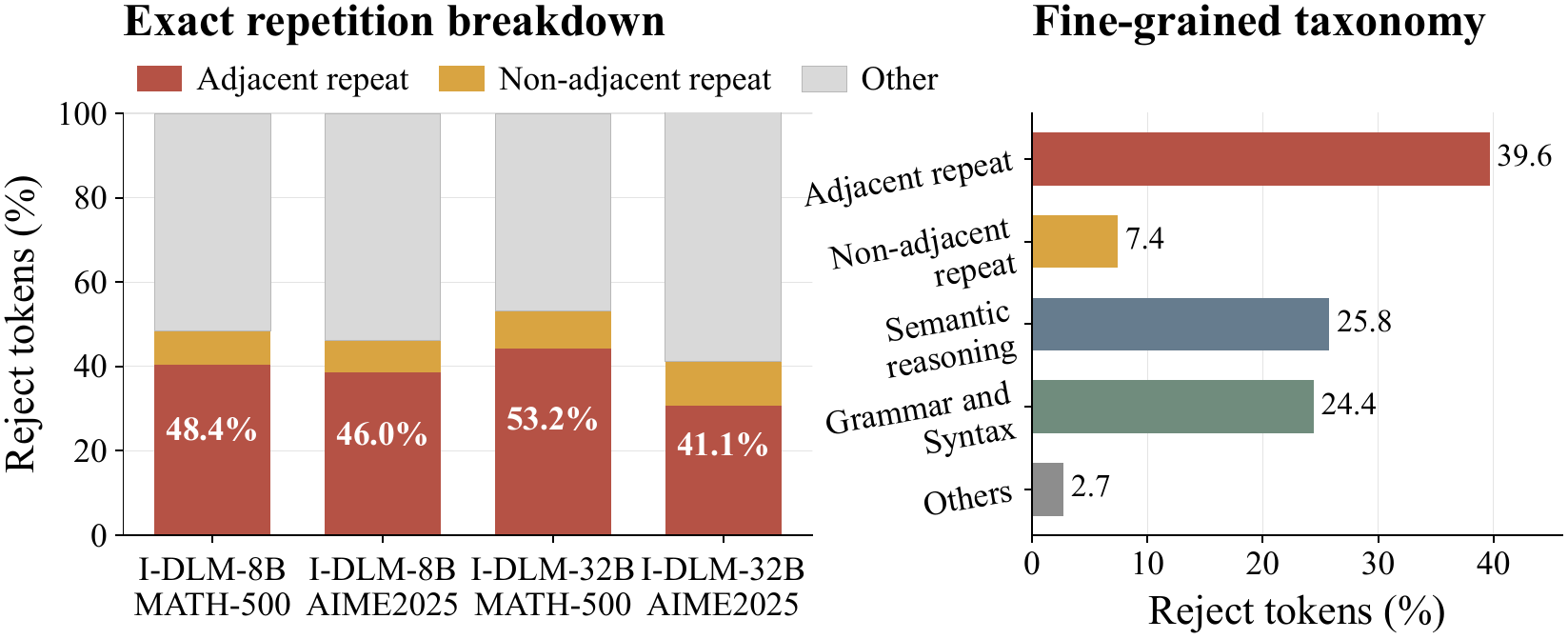}
        \vspace{-.5cm}
        \caption{
        \textbf{Empirical patterns of proposal rejection.} 
        Repetition accounts for a substantial fraction of rejected tokens, alongside cases involving semantic reasoning and grammar.
        }
        \label{fig:rejection_breakdown}
    \end{minipage}
    \vspace{-12pt}
\end{figure}

\subsection{Rejection Severely Limits Effective Throughput}

Block size measures \emph{nominal} parallelism, but realized decoding progress depends on how much of the proposed block survives verification. Consider a verifier that checks $M$ parallel proposals from left to right, and let $a_i$ denote the probability that position $i$ is accepted conditioned on the preceding prefix. The probability that verification reaches through position $k$ is $A_k=\prod_{i=1}^{k} a_i$. Hence, the expected accepted-prefix length is $
\mathbb{E}[L]
=\sum_{k=1}^{M} A_k$.
This simple relation captures the central inefficiency of verified parallel generation: acceptance compounds across positions. Even moderate per-position rejection can eliminate a large fraction of the useful parallelism available in a block.

The effect can be stronger when rejection also changes the number of model evaluations required by the decoder. For example, under the ISD execution rule~\citep{yu2026idlm}, a stride of $N$ yields
\begin{equation} 
\mathrm{TPF}_N(r) =\frac{2+\sum_{i=1}^{N-2}(1-r)^i}{2-(1-r)^{N-1}}
\end{equation}
where the denominator accounts for the additional forward-pass cost induced when a proposal block does not fully survive. Figure~\ref{fig:tpf_rejection} plots this relation for $N=8$. TPF drops from $8$ with no rejection to 4.09 at 10\% rejection, 3.29 at 15\%, and 2.77 at 20\%. Thus, seemingly modest changes in proposal reliability can translate into large differences in realized throughput.
Figure~\ref{fig:rejection_breakdown} further shows that these rejections are structured rather than uniform: exact repetition accounts for $41.1\%$--$53.2\%$ of rejected tokens, with adjacent repetition forming the largest fine-grained category. 

These observations motivate optimizing the acceptance profile rather than relying on block width or average token accuracy as proxies for decoding efficiency. Moreover, because progress depends on prefix survival, improvements at different positions need not have equal value. This motivates the decoder-aware credit assignment developed in Section~\ref{sec:method}.

\section{\name: Optimizing Effective Block Progress}
\label{sec:method}

\subsection{Overview}

In this paper, we consider the class of verified parallel decoders, in which a model proposes $M$ future tokens in parallel and a prefix-conditioned verifier determines how much of the proposal can be committed. This abstraction encompasses both DLM-drafter systems with a separate autoregressive verifier, such as DFlash~\citep{chen2026dflash} and D-Pace~\citep{wu2026dpace}, and self-verifying designs such as I-DLM~\citep{yu2026idlm}, where parallel proposal and causal verification are provided by the same model. BRISK-DLM operates on this shared proposal--verification structure.

This popular structure exposes two places where parallel prediction is misaligned with realized decoding progress. First, the proposal model is typically trained position-wise, although improving different positions can have very different effects on the amount of verified output obtained from a model evaluation. Second, the backbone computes all proposal positions before the token choices at earlier positions are known, even though candidates at a later position depend on this realized prefix.

\begin{figure}[t]
\centering
\includegraphics[width=.95\linewidth]{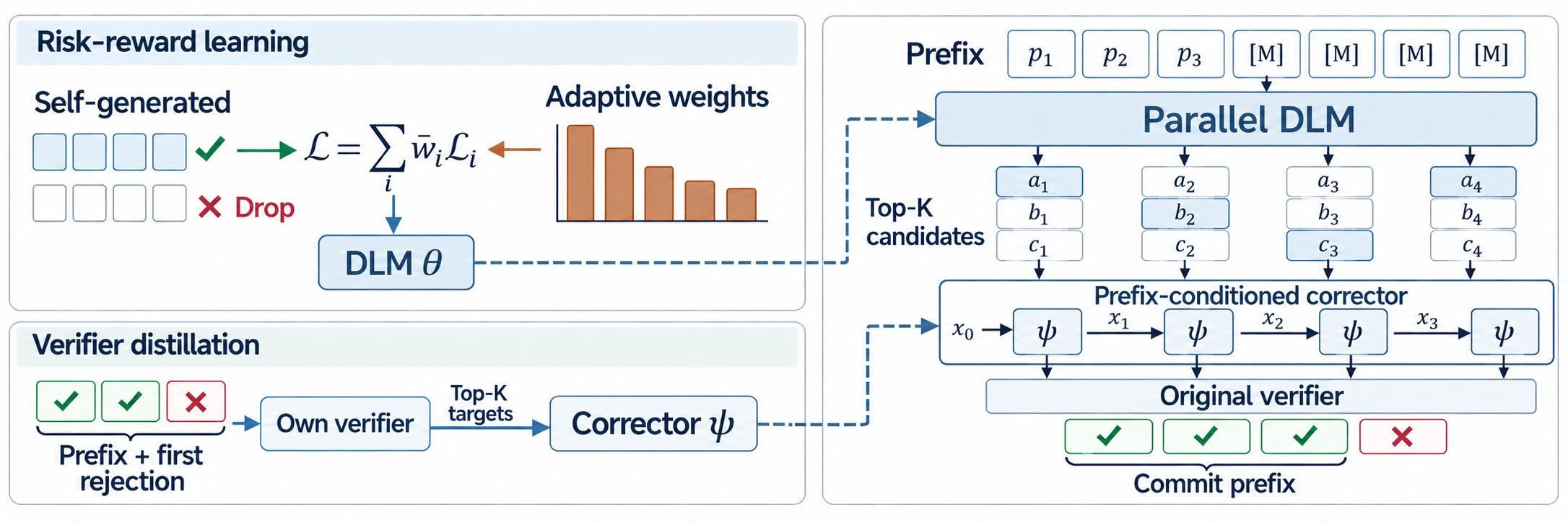}
\vspace{-.4cm}
\caption{\name overview. Left: risk--reward learning and
verifier distillation. Right: prefix-conditioned proposal
selection followed by verification.}
\label{fig:overview}
\end{figure}

BRISK-DLM addresses these two mismatches at their respective interfaces. As shown in Figure~\ref{fig:overview}, \emph{Risk-reward proposal learning} allocates model capacity according to each position's marginal contribution to verified progress (\S\ref{sec:proposal_learning}). \emph{Prefix-conditioned proposal selection} then reuses the parallel backbone representations while conditioning the final token choices on the prefix already selected within the block (\S\ref{sec:inference_head}). Together, \name improves candidate quality and assemble those candidates into lower-rejection blocks, while leaving the underlying DLM architecture, verifier, rejection correction, and stopping rule unchanged, and requiring no additional backbone forward pass.

\subsection{Risk–Reward Proposal Learning}
\label{sec:proposal_learning}

Inference-time selection can only choose among candidates supplied by the proposal model, so improving verified progress first requires improving the proposal distribution itself. BRISK-DLM aligns proposal learning with the model's realized decoding behavior along two dimensions: (i) \emph{which states it learns from} and (ii) \emph{where learning signal is allocated}. Fixed offline sequences do not reproduce the contexts and continuation patterns generated by the model itself, while uniform token-wise losses ignore that errors at different positions have different consequences for effective block progress. We address these issues as a single learning process: self-generated trajectories expose the proposal states and failure patterns encountered during decoding, and position-aware weights determine how the resulting training signal should be allocated.

Specifically, we sample multiple responses from the initial model checkpoint using the deployed decoding procedure and retain trajectories whose final answers are correct. Such filtering avoids reinforcing unsuccessful generations while preserving diverse solution trajectories, bringing the training distribution closer to inference. We analyzed the effect of data selection in Section~\ref{performance_breakdown}.

\noindent \textbf{Marginal verified-progress weighting.}
Within each retained trajectory, we adjust the contribution of every proposal position according to its effect on expected block progress. Improving an early position protects every subsequent proposal, whereas later positions become increasingly important once the preceding prefix is reliable. Hence, the relevant bottleneck changes as the model improves and cannot be captured by uniform weighting or a fixed positional schedule.

For a block of $M$ proposal positions, let $a_i$ be the probability that position $i$ is accepted conditioned on the preceding prefix, and define its survival probability as $A_k=\prod_{j=1}^{k}a_j$ with $A_0=1$. If $L$ denotes the accepted-prefix length, its distribution and the resulting decoding efficiency are
\begin{equation}
\begin{aligned}
    \Pr(L=k)
    &=
    \begin{cases}
        A_k(1-a_{k+1}), & 0\leq k<M,\\
        A_M, & k=M,
    \end{cases}\\[-1pt]
    \mathrm{TPF}(\mathbf{a})
    &=\frac{\sum_{k=0}^{M}\Pr(L=k)g_k}
            {\sum_{k=0}^{M}\Pr(L=k)c_k},
    \qquad
    \mathrm{FPT}(\mathbf{a})=\mathrm{TPF}(\mathbf{a})^{-1},
\end{aligned}
\label{eq:nonuniform_tpf}
\end{equation}
where $g_k$ and $c_k$ denote the number of committed tokens and the forward-pass cost when $L=k$, respectively. This general form accommodates both conventional speculative decoding, where a DLM drafter is verified by a separate AR target, and self-speculative DLM decoding such as I-DLM, where proposal and verification are carried out within the same model. For conventional speculative decoding and I-DLM with unit verification cost and one guaranteed committed token per forward pass, it reduces to$\mathrm{TPF}=1+\sum_{k=1}^{M}A_k$. In practice, we estimate \(a_i\) as one minus the empirical rejection rate at position \(i\) on self-generated training data, computed according to each architecture’s native verification rule and conditioned on verification reaching that position.

We measure the importance of position $i$ by its marginal effect on FPT under the log-odds parameterization $z_i=\log(a_i/(1-a_i))$. The raw importance, its mean-one normalization, and the resulting proposal objective are summarized as:
\begin{equation}
\begin{aligned}
    w_i^{\mathrm{raw}}
    &=-\frac{\partial\,\mathrm{FPT}(\mathbf{a})}{\partial z_i}
     =-\frac{\partial\,\mathrm{FPT}(\mathbf{a})}{\partial a_i}
       a_i(1-a_i),\\
    \bar w_i
    &=\frac{M w_i^{\mathrm{raw}}}
            {\sum_{j=1}^{M}w_j^{\mathrm{raw}}+\epsilon},\qquad
    \mathcal{L}_{\mathrm{BRISK}}=\frac{1}{M}\sum_{i=1}^{M}
      \bar w_i\mathcal{L}^{\mathrm{prop}}_i.
\end{aligned}
\label{eq:brisk_weighting}
\end{equation}
The normalized weights preserve the overall loss scale and are treated as fixed coefficients during backpropagation. Recomputing them from the current acceptance profile shifts emphasis toward the positions that most constrain block progress, without requiring an additional model forward pass.
Figure~\ref{fig:position_weight_evolution} illustrates this adaptive behavior during training: the weighting initially emphasizes early proposal positions, which are more likely to truncate the block, and gradually shifts toward later positions as the earlier prefix becomes more reliable.

\subsection{Prefix-Conditioned Proposal Selection} 
\label{sec:inference_head}

Risk--reward learning improves marginal proposals, but their parallel
representations are computed before the actual within-block choices
$x_{<i}$ are known. A candidate can therefore be plausible in isolation
but poorly matched to the selected prefix. At first rejections, the
verifier's top-ranked token is already in the base Top-16 in 96.97\% of
states (Table~\ref{tab:mechanism_summary}). This suggests that recomputing the backbone is often unnecessary, and what is missing is a lightweight mechanism that conditions candidate selection on the realized prefix.
Existing work has introduced lightweight causal heads to condition later
decisions on earlier token choices
\citep{huang2026domino,cheng2026dspark}. This restores missing prefix
information, but leaves a complementary question: \emph{how should such a
selector be supervised for verified progress?}

BRISK addresses both selection and supervision at this interface. The backbone still produces all candidate representations in parallel, while a lightweight corrector sequentially reranks only the existing Top-$K$ candidates using previously selected tokens. We train this selector by distilling the model's own verifier on policy-reached decisions through the first rejection. Thus, selected-token feedback provides the realized-prefix
information needed for refinement, while verifier distillation specifies
which candidate should be preferred at the decisions that determine usable
block progress.

\noindent \textbf{Rejection-aware candidate selection.}
For proposal position $i$, we reuse the frozen backbone representation $m_i$, proposal logits $\ell_i^q$, and cached context representation $h_c$. Let
$
    \mathcal C_i=\operatorname{TopK}(q_i,K)
$
denote the candidate set obtained from the normalized base proposal distribution $q_i$. We initialize a compact state as
$
    u_0=\operatorname{Init}_{\psi}(h_c,E(x_0))
$, 
where $x_0$ is the token immediately preceding the proposal block. Each selected token then conditions the next candidate decision:
\begin{equation}
u_i=f_\psi\left(
u_{i-1},h_c,m_i,E(x_{i-1}),\phi_i
\right),
\label{eq:inference_head_state}
\end{equation}
where $E$ is the frozen token embedding and $\phi_i$ contains position and proposal-score features. For each candidate $v\in\mathcal C_i$, the corrector predicts a gated low-rank residual on top of the base logit:
\begin{equation}
  \tilde\ell_{\psi,i}(v)
  =
  \ell_i^q(v)
  +
  g_{\psi,i}
  \left\langle W_su_i,W_cE_{\mathrm{LM}}(v)\right\rangle,
  \label{eq:candidate_residual}
\end{equation}
where $E_{\mathrm{LM}}(v)$ is the frozen output-vocabulary row,
and the projections and gate are learned. We select $x_i$ from refined logits and feed it into the next state update.
Let $\pi_i$ be the resulting proposal law
and $p_i$ the frozen causal verifier distribution under the valid prefix
and target sampling configuration.
Standard speculative sampling accepts with probability
$\min\{1,p_i(x_i)/\pi_i(x_i)\}$ and resamples from the normalized positive part
$[p_i-\pi_i]_+$ upon rejection, preserving $p_i$
\citep{leviathan2023fast,chen2023speculative}.
Top-$K$ restricts proposals only.

\noindent \textbf{Putting it into practice.}
The remaining question is how to train the corrector. Existing causal refiners commonly train on reference or teacher-forced
continuations \citep{huang2026domino,cheng2026dspark}. While this provides
valid causal supervision, it does not fully specify the learning problem
faced by our selector: at deployment, the corrector reranks the proposal
model's actual candidates under prefixes formed by earlier proposal choices,
and its utility is determined by the verification decisions that are
actually reached before the block stops. We therefore align supervision with both the prefix and the reached decisions that determine block progress.

For each recorded proposal block $z_{1:M}$ with $a$ accepted positions, we
supervise
$\mathcal R=\{1,\ldots,\min(a+1,M)\}$.
At each $i\in\mathcal R$, the model's own verifier distribution under the
recorded prefix $z_{<i}$ is restricted and renormalized to the available
candidate set $\mathcal C_i$. Accepted positions teach candidate choices
that preserve verified progress, while the first rejection supplies the
preference at the decision that terminates it. Table~\ref{tab:head_training_ablation} 
supports these design choices: richer verifier targets under teacher
forcing alone do not yield consistent serving gains, whereas extending
policy-reached supervision from accepted-only through the first rejection
improves verified progress and throughput. The full objective is given in
Appendix~\ref{app:head_implementation}.

\noindent \textbf{Efficient system integration.}
BRISK-DLM adds no backbone evaluation. The corrector reuses cached proposal representations and vocabulary rows, and we fuse candidate construction, recurrent updates, residual scoring, and selection into a GPU-resident execution path. The resulting block is then passed unchanged to the original verifier. 
Execution details are provided in the Appendix~\ref{app:head_system_optimization}.

\section{Evaluations}

\textbf{Models and benchmarks.}
We experiment on the 8B and 32B I-DLM checkpoints and use DAPO-Math-17K as training data. 
Mathematical reasoning is evaluated on GSM8K, MATH-500, MathBench, AIME~2024, and AIME~2025, while code generation is evaluated on HumanEval and MBPP. We use stride $N=8$ by default. We report accuracy for mathematical reasoning and pass@1 for code generation. Efficiency is measured by rejection rate, accepted-prefix length, TPF, and end-to-end throughput. Throughput is evaluated at concurrency $C\in\{1,2,4,8,16,32\}$ on NVIDIA H200 and A100 GPUs. Complete training and evaluation details are provided in Appendix~\ref{app:details}.

\textbf{Baselines.}
We compare against the following representative state-of-the-art systems:
\begin{denseitemize}

\item \emph{I-DLM}~\citep{yu2026idlm}: an introspective DLM that uses introspective strided decoding (ISD) to verify previously proposed tokens while advancing new tokens in the same forward pass. 

\item \emph{LLaDA~2} and \emph{SDAR}~\citep{bie2026llada21,cheng2025sdar}: state-of-the-art block-diffusion systems for parallel generation. LLaDA~2.1 accelerates decoding through token editing, while SDAR converts pretrained AR models into blockwise DLMs.

\item \emph{DFlash}~\citep{chen2026dflash}: a speculative decoding system that uses a lightweight block-diffusion model to draft multiple tokens in parallel for verification by an AR target. The official repository only released Qwen3-8B drafter for non-thinking generation; we evaluate the same drafter with target thinking enabled and disabled. The main quality and throughput comparisons use thinking mode to retain reasoning-capable operating points.

\item \emph{D-PACE}~\citep{wu2026dpace}: dynamically weights the DFlash drafter's training loss for speculative decoding with AR target model; we use its official training and the same target modes.

\end{denseitemize}

\begin{table*}[t]
\centering
\footnotesize
\renewcommand{\arraystretch}{1.1}
\setlength{\tabcolsep}{2.5pt}

\begin{adjustbox}{max width=\textwidth}
\begin{tabular}{
l
cccccccc
>{\columncolor{gray!15}}c
cc
>{\columncolor{gray!15}}c
c
}
\toprule
& \makecell{\textbf{LLaDA-2.1}\\\textbf{-mini}}
& \makecell{\textbf{LLaDA-2.0}\\\textbf{-flash}}
& \makecell{\textbf{LLaDA-2.1}\\\textbf{-flash}}
& \makecell{\textbf{SDAR}\\\textbf{8B}}
& \makecell{\textbf{SDAR}\\\textbf{30B-A3B}}
& \makecell{\textbf{DFlash}\\\textbf{OOD-think.}}
& \makecell{\textbf{DFlash}\\\textbf{native NT}}
& \makecell{\textbf{I-DLM}\\\textbf{8B}}
& \makecell{\textbf{BRISK-DLM}\\\textbf{8B}}
& \makecell{\textbf{Qwen3}\\\textbf{8B}}
& \makecell{\textbf{I-DLM}\\\textbf{32B}}
& \makecell{\textbf{BRISK-DLM}\\\textbf{32B}}
& \makecell{\textbf{Qwen3}\\\textbf{32B}} \\

\textit{Params}
& \textit{16B} & \textit{100B} & \textit{100B}
& \textit{8B} & \textit{30B}
& \textit{8B} & \textit{8B}
& \textit{8B} & \textit{8B} & \textit{8B}
& \textit{32B} & \textit{32B} & \textit{32B} \\

\midrule
GSM8K
& 89.0 & \textbf{96.1} & 92.9
& 91.7 & 91.4 & 93.0 & 81.7
& 92.9 & 92.3 & 91.8
& 94.9 & \underline{95.4} & 94.7 \\

MATH-500
& 85.0 & 97.6 & 91.0
& 78.6 & 77.8 & 96.6 & 85.2
& 96.6 & 96.8 & \textbf{98.0}
& 97.6 & 96.0 & \underline{97.8} \\

MathBench
& 84.2 & 92.0 & 90.8
& 76.9 & 79.3 & 92.8 & 83.6
& 89.0 & 89.0 & 92.8
& \textbf{95.6} & 92.8 & \underline{95.5} \\

AIME-24
& 43.3 & 63.3 & 63.3
& 10.0 & 16.7 & 76.7 & 26.7
& 73.3 & \underline{80.0} & 76.7
& \textbf{83.3} & 76.7 & 76.7 \\

AIME-25
& 43.3 & 60.0 & 63.3
& 10.0 & 10.8 & 63.3 & 23.3
& 53.3 & 56.7 & \underline{66.7}
& \textbf{80.0} & \textbf{80.0} & \textbf{80.0} \\

\addlinespace[3pt]
\midrule
\addlinespace[3pt]

HumanEval
& 86.0 & \textbf{94.5} & 90.2
& 78.7 & 42.1 & 75.0 & 81.1
& 93.3 & 90.8 & 90.9
& \underline{94.3} & 93.3 & \textbf{94.5} \\

sanitized MBPP
& 82.1 & 88.3 & 88.4
& 72.0 & 32.3 & 75.1 & 75.1
& 92.2 & 88.9 & 89.9
& \underline{92.7} & \textbf{94.2} & 92.2 \\

\bottomrule
\end{tabular}
\end{adjustbox}

\caption{Performance on mathematical-reasoning and coding benchmarks.
Bold and underlined values indicate the best and second-best results,
respectively.}
\label{tab:quality}
\end{table*}

\subsection{Main Results}

\begin{figure}[t]
\centering
\vspace{-.3cm}
\includegraphics[width=\linewidth]{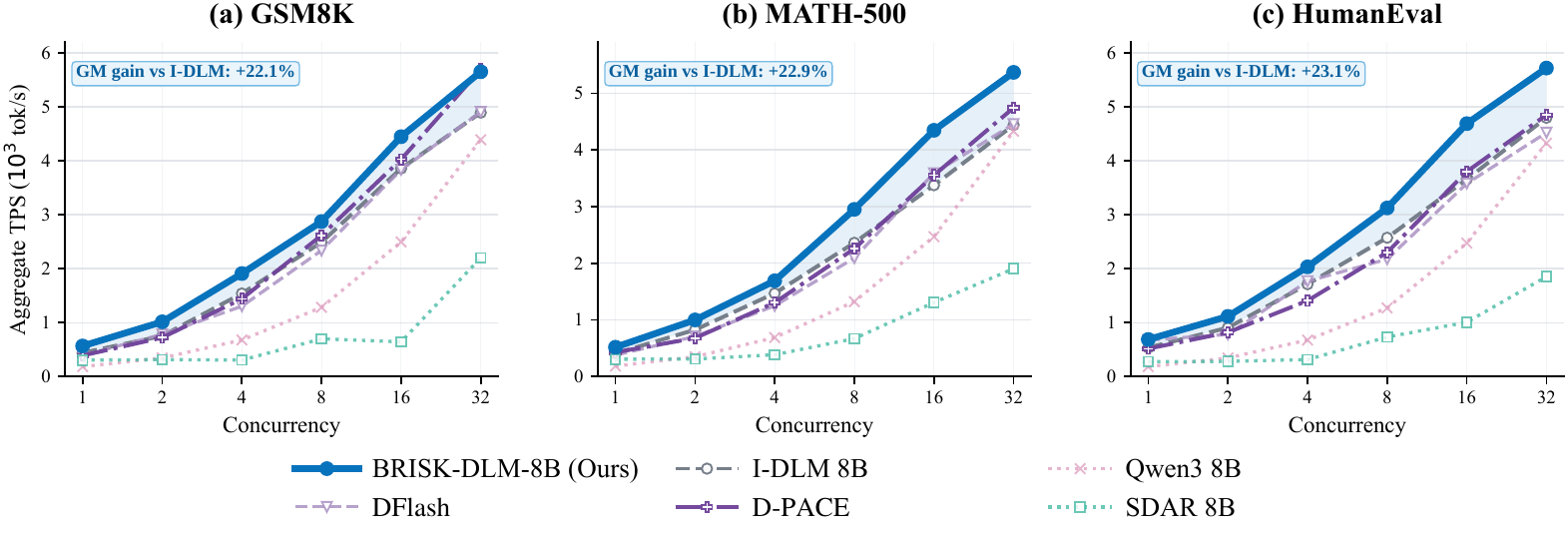}\par
\includegraphics[width=\linewidth]{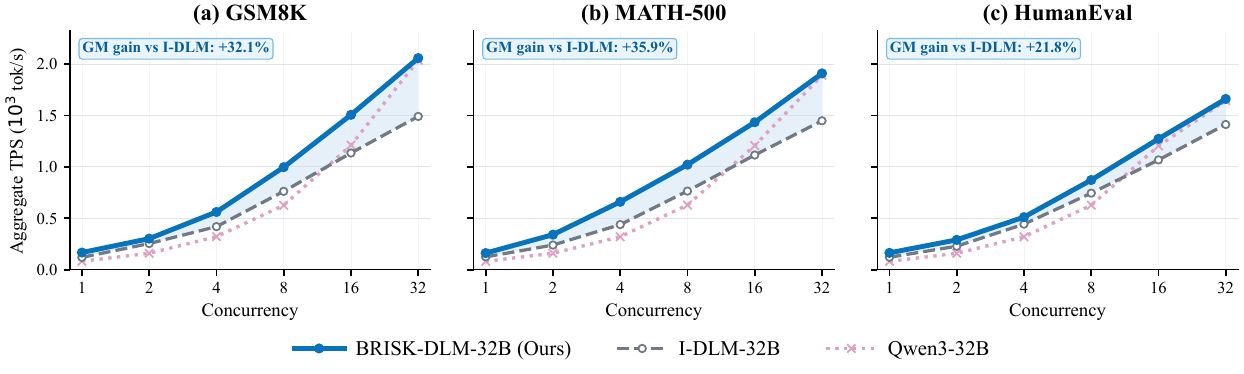}
\vspace{-.8cm}
\caption{\name improves throughput across model sizes, tasks, and concurrency levels.}
\label{fig:h200_throughput}\label{fig:h200_32b_throughput}
\vspace{-10pt}
\end{figure}

\textbf{\name improves end-to-end throughput across scales.}
Figure~\ref{fig:h200_throughput} reports serving throughput under matched H200 allocations. \name outperforms the same-scale I-DLM baseline in all 18 workload--concurrency settings at both model scales. At 8B, the geometric-mean gain is 22.7\%, with 15.1\%--34.7\% pointwise gains; task-wise gains are 22.1\% on GSM8K, 22.9\% on MATH-500, and 23.1\% on HumanEval. At 32B, the geometric-mean gain increases to 29.8\%, with 15.4\%--50.4\% pointwise gains; the corresponding task-wise gains are 32.1\%, 35.9\%, and 21.8\%. The larger gain at 32B is accompanied by a greater improvement in verified tokens per forward (37.4\% vs. 31.1\%), suggesting that increased verified progress contributes to the stronger throughput improvement. These results measure the complete serving path, including the execution overhead of the prefix-conditioned corrector.

\name also compares favorably with external systems. At 8B, relative to the thinking configurations of DFlash and D-PACE, \name achieves 1.29$\times$ and 1.27$\times$ geometric-mean TPS, respectively. Relative to Qwen3, it achieves 2.28$\times$ the TPS at 8B and 1.52$\times$ the aggregate TPS at 32B. 
Complete H200 results, including native NT baselines, are reported in Appendix (Table~\ref{tab:h200_full}).

\textbf{\name preserves strong task quality.}
Table~\ref{tab:quality} shows that \name remains broadly competitive with state-of-the-art, same-scale DLMs across reasoning and code-generation benchmarks. At 8B, average accuracy across the five mathematical reasoning benchmarks increases from 81.0 to 83.0. \name reaches 96.8 on MATH-500 and 56.7 on AIME~2025, the latter improving by 3.4 points over the matched I-DLM baseline, while performance on the remaining tasks changes only moderately. At 32B, results vary across tasks but remain strong overall: \name attains 80.0 on AIME~2025 and 93.3 on HumanEval while improving both GSM8K and MBPP.

\textbf{\name delivers more effective progress per model evaluation.}
To isolate the algorithmic source of the throughput gains, we measure committed tokens per forward (TPF) (Table~\ref{tab:matched_tpf}). On MATH-500, GSM8K, and HumanEval, \name increases 8B TPF from 2.79, 2.90, and 3.71 to 3.50, 4.26, and 4.53, respectively. At 32B, TPF increases from 2.05, 1.97, and 1.99 to 2.73, 2.81, and 2.71. These correspond to geometric-mean gains of 31.1\% at 8B and 37.4\% at 32B over matched I-DLM. 
The larger gains in TPF than in wall-clock throughput are expected: BRISK-DLM commits more output per expensive model evaluation, while the prefix-conditioned corrector introduces additional lightweight work. The end-to-end gains above show that, after this overhead is included, the increase in effective block progress still translates into substantial serving acceleration.

\subsection{Performance Breakdown and Ablation Studies}
\textbf{Performance Breakdown.}
\label{performance_breakdown}
Table~\ref{tab:ablation_breakdown} separates the contributions of self-generated data, dynamic position weighting, and prefix-conditioned selection. 
The results show that data selection and position weighting address different failure modes. With dynamic weighting fixed, training on self-generated trajectories yields higher TPF and throughput than training only on correctness-filtered trajectories, but reduces accuracy by 4.8 points. Correctness filtering recovers task quality at only a 2.5\% throughput cost. Replacing self-generated trajectories with open-source data is substantially worse: accuracy drops by 12.8 points, rejection rises from 14.03\% to 19.02\%, and throughput decreases by 22.3\%. This highlights the need for training on states induced by the decoding policy. 

Holding correctness-filtered self-generated data fixed, dynamic weighting reduces rejection by 1.52 points relative to uniform weighting and improves TPF and throughput by 7.0\% and 16.3\%, respectively, without reducing accuracy. Thus, policy-matched data determines which states the model learns from, while marginal-progress weighting determines where learning capacity is spent within those states. 
 The prefix-conditioned corrector provides a further inference-time gain on top of the trained proposal model. With the backbone checkpoint fixed, enabling the corrector reduces rejection from 14.03\% to 11.10\%, increases TPF from 3.35 to 3.63 (+8.4\%), and raises throughput from 1847.2 to 1948.8 tok/s (+5.5\%).

\begin{table*}[t]
\centering
\scriptsize

\begin{minipage}[t]{0.455\textwidth}
\vspace{0pt}
\centering
\setlength{\tabcolsep}{2.2pt}
\renewcommand{\arraystretch}{1.12}

\begin{tabularx}{\linewidth}{@{}Xrrrr@{}}
\toprule
Configuration & GSM8K & MATH-500 & HumanEval & Avg. gain \\
\midrule
I-DLM-8B
  & 2.897 & 2.794 & 3.710 & -- \\
BRISK-DLM
  & 4.264 & 3.497 & 4.534 & +31.1\% \\
\midrule
I-DLM-32B
  & 1.967 & 2.048 & 1.990 & -- \\
BRISK-DLM
  & 2.805 & 2.734 & 2.713 & +37.4\% \\

\bottomrule
\end{tabularx}

\caption{TPF for same-scale I-DLM and BRISK pairs.}
\label{tab:matched_tpf}
\end{minipage}
\hfill
\begin{minipage}[t]{0.525\textwidth}
\vspace{0pt}
\centering
\setlength{\tabcolsep}{2.0pt}
\renewcommand{\arraystretch}{1.12}

\begin{tabularx}{\linewidth}{
    @{}X
    >{\centering\arraybackslash}p{0.035\linewidth}
    >{\centering\arraybackslash}p{0.055\linewidth}
    rrrrr@{}
}
\toprule
Training data & W. & Head & Acc. (\%) & Reject (\%) & TPF & TPS \\
\midrule
Open-source
  & D & Off & 83.4 & 19.02 & 2.91 & 1436.2 \\
Self-gen. (all)
  & D & Off & 91.4 & 13.87 & 3.41 & 1893.6 \\
Self-gen. (corr.)
  & U & Off & 96.0 & 15.55 & 3.13 & 1589.0 \\
Self-gen. (corr.)
  & D & Off & 96.2 & 14.03 & 3.35 & 1847.2 \\
\textbf{Self-gen. (corr.)}
  & \textbf{D}
  & \textbf{On}
  & \textbf{96.2}
  & \textbf{11.10}
  & \textbf{3.63}
  & \textbf{1948.8} \\
\bottomrule
\end{tabularx}

\caption{Training choices and head impact (MATH-500).
U and D denote uniform and dynamic weights.}
\label{tab:ablation_breakdown}
\end{minipage}
\vspace{-12pt}
\end{table*}

\textbf{Cross-backbone transfer.}
\label{sec:cross_backbone_transfer}
To test whether proposal learning is specific to I-DLM, we apply the BRISK training recipe to the DFlash drafter while keeping the Qwen3-8B target, serving path, and numeric path fixed. Under native non-thinking GSM8K requests, Table~\ref{tab:dflash_training_transfer} shows a 18.3\% geometric-mean TPS gain, with improvements at all six concurrency levels. 
\begin{table}[t]
\centering
\small
\setlength{\tabcolsep}{3.4pt}
\renewcommand{\arraystretch}{1.08}
\begin{tabular}{@{}lrrrrrrr@{}}
\toprule
Drafter & $C=1$ & $C=2$ & $C=4$ & $C=8$ & $C=16$ & $C=32$ & GM gain \\
\midrule
Original DFlash & 878.4 & 1612.5 & 2784.2 & 4531.7 & 6611.0 & 10046.0 & -- \\
BRISK-DFlash & \textbf{925.8} & \textbf{2217.1} & \textbf{3568.7} & \textbf{5755.6} & \textbf{7141.1} & \textbf{10788.0} & \textbf{+18.3\%} \\
\bottomrule
\end{tabular}
\caption{Training-recipe transfer to DFlash on GSM8K. Aggregate TPS (model tokens/s) on one H200, with exactly 2,048 generated model tokens per request.}
\label{tab:dflash_training_transfer}
\vspace{-12pt}
\end{table}

\textbf{Residual proposal errors and repair.}
\label{sec:mechanism_main}
Even after proposal training, many first rejections occur despite a
verifier-preferred alternative remaining available within the restricted
candidate support: the verifier's top-ranked token is already contained
in the base Top-16 at \MechTopOneCoverage{} of factual first-breaker
states. Under same-state replay, prefix-conditioned refinement rescues
\MechRescue{} decisions per 100 first breakers; separately, it introduces
\MechRegression{} rejections per 100 accepted positions. Together, the diagnostics support the view that a substantial part
of the residual error lies in selecting among available candidates under
the realized prefix. Appendix~\ref{app:mechanism_analysis} provides the
full analysis, including replay definitions, position-wise
intervention and rescue, and whole-block verifier-likelihood outcomes.

\textbf{What should supervise the candidate selector?}
\label{sec:head_training_ablation}
Table~\ref{tab:head_training_ablation} asks two questions:
\emph{under which prefix should the selector learn verifier preferences},
and \emph{which reached decisions should be supervised}. 
We first ask whether richer targets are sufficient under teacher forcing.
The teacher-forced controls reflect reference-conditioned supervision used
by representative causal refiners
\citep{huang2026domino,cheng2026dspark}, without reimplementing those
systems. Replacing hard token labels with soft verifier targets does not
yield consistent serving gains, showing that target richness alone is
insufficient. BRISK instead trains on policy-reached candidate decisions
with verifier targets, and achieves 780.3 and 1198.7 TPS at $C=8$ and
$16$, outperforming both teacher-forced controls at these loads.

We next ask how far supervision should extend along a policy-reached
trajectory. Accepted-only targets teach the selector how to preserve an already
successful prefix, but omit the first reached decision that terminates
usable block progress. BRISK retains this position and distills the
verifier's preference over its available candidates. Relative to accepted-only supervision, BRISK improves TPS from 744.4 to 780.3 at C=8 and from 1123.1 to 1198.7 at C=16.

\begin{table}[t]
\centering
\begin{minipage}[t]{0.64\linewidth}
\centering\vspace{0pt}
\footnotesize\setlength{\tabcolsep}{2.0pt}\renewcommand{\arraystretch}{1.08}

\begin{tabular}{@{}lrrr@{}}
\toprule
& \multicolumn{3}{c}{Mean TPS / request TPF $\uparrow$} \\
\cmidrule(lr){2-4}
Supervision & $C=1$ & $C=8$ & $C=16$ \\
\midrule
No corrector & 91.6/3.42 & 704.8/3.46 & 1055.1/3.38 \\
\makecell[l]{TF token\\labels} & 96.1/3.68 & 753.6/3.74 & 1115.8/3.68 \\
\makecell[l]{TF verifier\\targets} & 95.4/3.66 & 743.1/3.71 & 1152.6/3.71 \\
\makecell[l]{Policy-reached,\\accepted only} & 98.1/\textbf{3.76} & 744.4/3.74 & 1123.1/3.73 \\
\makecell[l]{\textbf{BRISK: through}\\\textbf{first rejection}} & \textbf{98.2/3.76} & \textbf{780.3/3.76} & \textbf{1198.7/3.79} \\
\bottomrule
\end{tabular}

\caption{Selector supervision at $N=8$ on A100. Cells show mean TPS / mean request TPF. TF denotes teacher forcing.}
\label{tab:head_training_ablation}
\end{minipage}\hfill
\begin{minipage}[t]{0.34\linewidth}
\centering\vspace{0pt}
\footnotesize\setlength{\tabcolsep}{2.0pt}\renewcommand{\arraystretch}{1.15}
\begin{tabular}{@{}rrr@{}}
\toprule
& First breakers & Natural EOS\\
\cmidrule(lr){2-2}\cmidrule(l){3-3}
$K$ & \makecell{Coverage/\\mass\,(\%)} & \makecell{Prefix/\\TPF}\\
\midrule
4 & 80.12/80.03 & 4.83/4.12\\
8 & 91.50/91.40 & 4.84/4.12\\
16 & 96.56/96.49 & \textbf{5.04/4.36}\\
32 & \textbf{98.71/98.67} & 4.79/4.07\\
\bottomrule
\end{tabular}

\caption{Top-$K$ sensitivity on MATH-500. Coverage and mass describe candidate availability, not an oracle rescue rate.}
\label{tab:topk_main}
\end{minipage}
\vspace{-12pt}
\end{table}

\textbf{Top-$K$ sensitivity analysis.}
\label{sec:topk_analysis}
We vary $K\in{4,8,16,32}$ with the 8B BRISK backbone fixed. At head-off first rejections, $K=16$ already contains the verifier's top-1 choice in 96.56\% of states and captures 96.49\% of its probability mass (Table~\ref{tab:topk_main}); increasing to $K=32$ adds only about $2.2$ points, while ranks 17--32 are selected in just 0.07\% of cases. On MATH-500, $K=16$ achieves the highest measured accepted-prefix length and TPF, and the highest A100 throughput at all three evaluated concurrency levels, $C\in\{1,8,16\}$. We use $K=16$ as the default.

\textbf{Impact of Stride Size.} 
We sweep $N\in{4,8,16}$ on MATH-500 to test whether \name's gains depend on the default stride. Across all strides and concurrency levels, \name consistently outperforms matched I-DLM, with geometric-mean TPS gains of 14.7\%, 22.9\%, and 20.2\% for $N=4,8,16$, respectively (Appendix Table~\ref{tab:stride_sensitivity}). Thus, BRISK's benefit is robust to proposal width.

\textbf{System optimization.}
For the same correcter checkpoint, a reference implementation
improves TPF but reduces end-to-end TPS relative to head-off decoding.
Optimized execution instead delivers an 8.2--9.5\% TPS gain over
head-off decoding on A100 across $C\in\{1,8,32\}$
(Appendix~Table~\ref{tab:head_system_ablation}).

\section{Conclusion}

BRISK-DLM aligns parallel proposal learning and selection with the prefix-dependent progress that ultimately determines decoding efficiency. It combines correctness-filtered self-generated training, marginal-progress weighting, and prefix-conditioned selection while preserving efficient parallel backbone computation. Across AR--DLM and DLM decoding settings, BRISK-DLM consistently improves generation efficiency by 22.7\% at 8B and 29.8\% at 32B, while maintaining task quality.

\newpage

\newpage

\bibliography{2027_conference}
\bibliographystyle{2027_conference}

\clearpage
\appendix

\section{Detailed Related Work}
\label{related}
\paragraph{Diffusion language models.}
Diffusion language models (DLMs) generate text by iteratively recovering tokens from partially masked states, offering a route to parallel generation beyond left-to-right autoregression. Early discrete diffusion formulations established principled denoising objectives in discrete spaces \citep{austin2021d3pm}, while Diffusion-LM \citep{li2022diffusionlm}, masked diffusion language models \citep{sahoo2024mdlm}, and SEDD \citep{lou2024sedd} developed these ideas for language generation. Recent systems have substantially expanded the scale and decoding design space of DLMs, including LLaDA \citep{nie2025llada}, Block Diffusion \citep{arriola2025blockdiffusion}, DREAM \citep{ye2025dream}, SDAR \citep{cheng2025sdar}, and LLaDA~2 \citep{bie2025llada20,bie2026llada21}. A complementary line converts pretrained autoregressive models into DLMs to reduce training cost \citep{gong2024scaling,deschenaux2024selfdistillation,fu2025efficientdlm,su2026opdlm}. In contrast to iterative denoising methods, I-DLM uses introspective strided decoding (ISD) to generate new tokens while verifying previously proposed tokens in the same forward pass \citep{yu2026idlm}. BRISK-DLM builds on this setting: our focus is not a new DLM architecture or a new denoising schedule, but improving the proposal acceptance profile that determines how much ISD parallelism is realized in practice.

\paragraph{Parallel drafting and acceptance-aware training.}
Speculative decoding accelerates autoregressive generation by verifying a fast model's draft with a target model while preserving the target distribution \citep{leviathan2023fast,chen2023speculative}. Subsequent work improves drafting through multiple heads \citep{cai2024medusa}, feature-level proposals \citep{li2024eagle,li2025eagle3}, tree verification \citep{miao2023specinfer}, and multi-token prediction \citep{gloeckle2024mtp}. Parallel block drafters such as DFlash \citep{chen2026dflash} make accepted-prefix length especially important because all draft positions are proposed jointly. 
Most closely related, D-PACE uses prefix-dependent acceptance to derive dynamic position weights\citep{wu2026dpace}. BRISK shares the observation that proposal positions have unequal effects on accepted progress, but differs in three respects. First, our objective accounts for forward-pass cost in addition to accepted progress. Second, we treat prefix dependence as an inference-time selection problem as well as a training-time weighting problem, using a lightweight corrector to rerank parallel candidates under the realized proposal prefix. Third, BRISK targets the proposal–verification interface internal to the DLM itself rather than a separate diffusion-drafter/AR-target pipeline.
GRIFFIN instead mitigates training--decoding token misalignment through token alignment and loss masking \citep{hu2025griffin}. These methods address a separate drafter--target configuration. Our work studies the proposal and verification paths internal to DLM, and couples acceptance-aware optimization with correct self-generated trajectories and proposal-side block consistency.

\section{Mechanism Analysis}
\label{app:mechanism_analysis}
\paragraph{Setup.}
We analyze 100 GSM8K and 100 MATH-500 requests at $N=8$, with seven speculative positions and Top-16 support. The factual head-off policy uses the Online+Dynamic checkpoint; the head-enabled policy uses the same backbone. Local substitutions are evaluated at factual head-off states. Whole-block replay starts both selectors from shared producer states. These diagnostic runs are separate from throughput measurement; Table~\ref{tab:mechanism_summary} summarizes the main findings.

\subsection{First-breaker repair and block compatibility}
\label{app:mech_support_repair}
A first breaker is the first speculative token rejected by ISD. The analysis contains 62,079 factual first breakers. Candidate coverage in Table~\ref{tab:mechanism_summary} asks whether the clean verifier's full-vocabulary top-ranked token is in the base Top-16. Coverage describes available support; it is not an oracle acceptance rate.

\begin{figure}[t]
\centering
\includegraphics[width=0.94\linewidth]{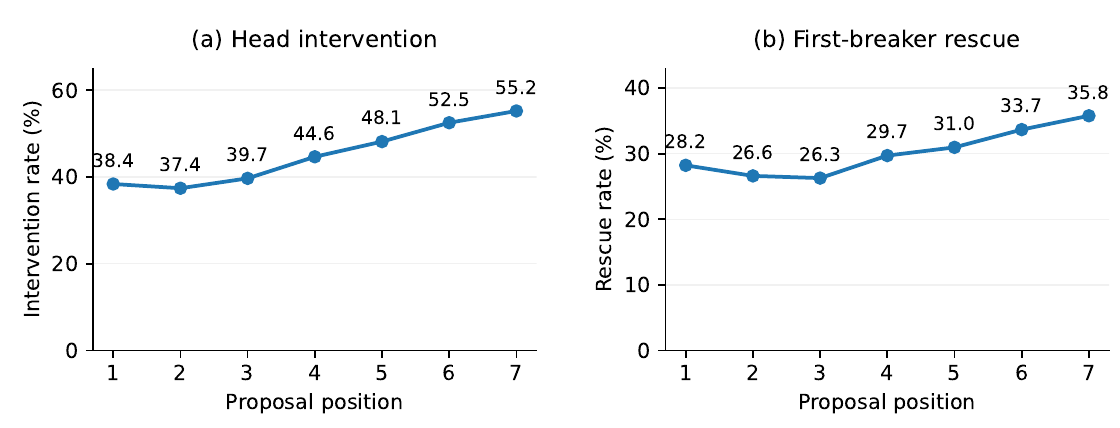}
\caption{\textbf{First-breaker repair by proposal position.}
(a) Head intervention and (b) same-draw rescue, each as a fraction of factual
first breakers at that position. Intervention rises from 38.4\% at position 1
to 55.2\% at position 7, while rescue rises from 28.2\% to 35.8\%.
The overall upward trend suggests that later proposals offer more opportunity
for prefix-conditioned correction. The rescue curve shows that the head's
additional interventions translate into more accepted local decisions,
rather than merely changing token choices.}
\label{fig:mech_first_breaker_position}
\end{figure}

\paragraph{Same-state replay.}
We fix the factual prefix, backbone representations, Top-16 support, full-vocabulary-normalized base $q_i$, clean verifier $p_i$, and acceptance draw $u_i$, changing only the token selected at the intervention position. The evaluated diagnostic rule is
\begin{equation}
\alpha_i(x)=\min\{1,p_i(x)/q_i(x)\},\qquad J_i(x)=\mathbf1[u_i<\alpha_i(x)].
\label{eq:mech_same_draw}
\end{equation}
A rescue is $J_i(x^{\mathrm{base}})=0$, $J_i(x^{\mathrm{head}})=1$ at a factual first breaker. A regression is the reverse outcome at a factual accepted position. Their denominators differ, so subtracting their displayed conditional rates is not a net rescue rate. Table~\ref{tab:mechanism_summary} adds the intervention rate; Figure~\ref{fig:mech_first_breaker_position} provides position-wise results. Events from the same request are correlated, and local summary rates use equal-request averaging rather than treating every event as an independent request.

Rescue and regression quantify candidate substitutions under this fixed-base-$q$ diagnostic rule. A gain in $p_i(x)/q_i(x)$ need not by itself imply a gain in $p_i(x)$; clean-verifier likelihood is therefore assessed separately.

\paragraph{Whole-block replay.}
\label{app:mech_block}
The head selects all seven positions sequentially. Base and head blocks are then scored in separate clean-verifier evaluations from their shared producer prefix $c$:
\begin{equation}
\mathcal L(b\mid c)=\sum_{i=1}^{7}\log p(x_i\mid c,x_{<i}),\qquad
\Delta\mathcal L=\mathcal L(b^{\mathrm{head}}\mid c)-\mathcal L(b^{\mathrm{base}}\mid c).
\label{eq:mech_block_logp}
\end{equation}
Each block is scored under its own selected predecessor path. The mean gain uses equal-request averaging; outcome fractions pool 109,197 paired blocks. Improved, tied, and regressed blocks are all retained in Table~\ref{tab:mechanism_summary}. Equal likelihood does not establish token identity, and higher likelihood does not establish semantic or final-answer correctness.

\begin{table}[t]
\centering
\small
\setlength{\tabcolsep}{5pt}
\caption{
Mechanism diagnostics for residual selection errors.
Candidate-availability and local-substitution statistics are
equal-request averages over 100 GSM8K and 100 MATH-500 requests.
}
\label{tab:mechanism_summary}
\begin{tabular}{@{}llr@{}}
\toprule
Quantity & Population / unit & Value \\
\midrule
\multicolumn{3}{l}{\textit{Candidate availability and same-state repair}} \\
Verifier top-1 in base-$q$ Top-16
    & First breakers & \MechTopOneCoverage{} \\
Corrector changes selected token
    & First breakers & \MechIntervention{} \\
Corrector rescues rejection
    & Per 100 first breakers & \MechRescue{} \\
Corrector introduces rejection
    & Per 100 accepted positions & \MechRegression{} \\
\midrule
\multicolumn{3}{l}{\textit{Whole-block verifier preference}} \\
Mean $\Delta\mathcal L$
    & Nats per seven-token block & +\MechDeltaBlockLogP{} \\
Refined block has higher likelihood
    & Paired blocks & \MechBlockPositive{} \\
Equal recorded likelihood
    & Paired blocks & \MechBlockTied{} \\
Base block has higher likelihood
    & Paired blocks & \MechBlockNegative{} \\
\bottomrule
\end{tabular}
\end{table}

\subsection{Local repetition}
\label{app:mech_collision}
\begin{figure}[t]
\centering
\includegraphics[width=0.48\linewidth]{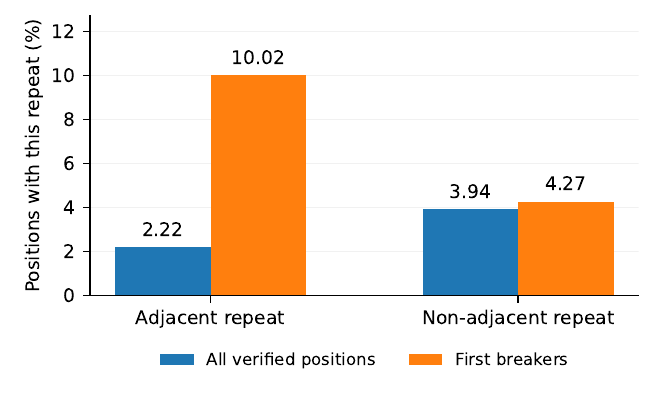}
\caption{\textbf{Repeat prevalence among verified positions and first breakers.}
Rates are equal-request averages.}
\label{fig:mech_collision}
\end{figure}

Figure~\ref{fig:mech_collision} compares equal-request repeat prevalence over verified positions and first breakers. Adjacent repetition increases from \MechAdjacentIncidence{} to \MechAdjacentBreakerShare{}, whereas non-adjacent repetition changes from \MechNonadjacentIncidence{} to \MechNonadjacentBreakerShare{}. This is an association in the trained head-off policy, not evidence that repeated tokens are necessarily invalid or are the primary cause of rejection.

Under the same-draw diagnostic rule, the head rescues \MechAdjacentHeadRescue{} of adjacent-repeat first breakers and \MechNonadjacentHeadRescue{} of non-adjacent-repeat first breakers. Paired whole-block replay shows adjacent repeats decreasing from 0.404 to 0.101 tokens per block; non-adjacent repeats change from 0.348 to 0.364. The selective change is consistent with prefix-dependent refinement rather than uniform repetition suppression. These paired-block counts are not the same quantity as the rejection-conditioned percentages in the earlier baseline taxonomy.

\section{Implementation Details}
\label{app:details}

\subsection{Brisk Backbone}
\label{app:training_details}

We use DAPO-Math-17K as the source of training prompts and reference
answers. For each prompt, we generate candidate responses once from
the initial I-DLM checkpoint using the deployed decoding procedure.
We generate the training data offline rather than continuously
regenerating it during training to mitigate potential overfitting.
We retain only trajectories whose final answers match the reference
under normalized answer matching, reducing the risk of reinforcing
incorrect reasoning paths. The resulting correctness-filtered
self-generated trajectories are used for continued training.
All runs use the original I-DLM masking construction with stride
$N=8$, yielding $M=N-1=7$ proposal positions. The position-risk
weights are applied to these seven proposal losses according to
Eq.~\eqref{eq:brisk_weighting}. Controlled ablations use the same
initialization, training-token budget, and optimization schedule.

We estimate $a_i$ from position-wise rejection statistics on the
self-generated training data, following the native introspective
strided decoding (ISD) verification rule.
At position $i$, the rejection rate is the number of proposals
rejected at that position divided by the number of blocks in which
verification reaches it; $a_i$ is one minus this rate.
Positions after the first rejection are excluded from the statistics
because they are not reached by verification.

Using this acceptance profile, the general renewal-reward
formulation in Eq.~\eqref{eq:nonuniform_tpf} specializes to
\begin{equation}
    \mathrm{TPF}(\mathbf{a})
    =
    \frac{
        2 + \sum_{k=1}^{M-1} A_k
    }{
        2 - A_M
    },
    \qquad M=N-1.
    \label{eq:idlm_tpf_instantiation}
\end{equation}
This corresponds to $g_k=2+\min(k,M-1)$, with $c_k=2$ for $k<M$
and $c_M=1$.
In all experiments, we use Eq.~\eqref{eq:idlm_tpf_instantiation}
to compute $\mathrm{FPT}(\mathbf{a})=1/\mathrm{TPF}(\mathbf{a})$,
from which the raw position sensitivities and normalized weights
are obtained using Eq.~\eqref{eq:brisk_weighting}.
The normalized weights are treated as fixed coefficients during
backpropagation.
Table~\ref{tab:implementation_details} summarizes the remaining
training and inference settings.

\begin{table}[h]
\centering
\caption{Training and inference configurations.}
\label{tab:implementation_details}
\small
\setlength{\tabcolsep}{4pt}
\renewcommand{\arraystretch}{0.92}
\begin{tabular}{lcc}
\toprule
Setting & 8B & 32B \\
\midrule
Training data & All valid & All valid \\
Optimizer / LR & AdamW / $2{\times}10^{-5}$ & AdamW / $10^{-5}$ \\
Global batch size & 32 & 32 \\
Sequence length & 4096 & 8192 \\
Training epochs & 5 & 5 \\
Training GPUs & $8{\times}$H20 & $8{\times}$H20 \\
Tensor parallelism & 1 & 4 \\
Precision & BF16 & BF16 \\
CUDA / PyTorch & 12.1 / 2.5.1 & 12.1 / 2.5.1 \\
SGLang / Triton & 0.4.6 / 3.1.0 & 0.4.6 / 3.1.0 \\
\bottomrule
\end{tabular}
\end{table}

\subsection{Adaptive Position-Weight Dynamics}
\begin{figure}[h]
    \centering
    \includegraphics[width=\linewidth]{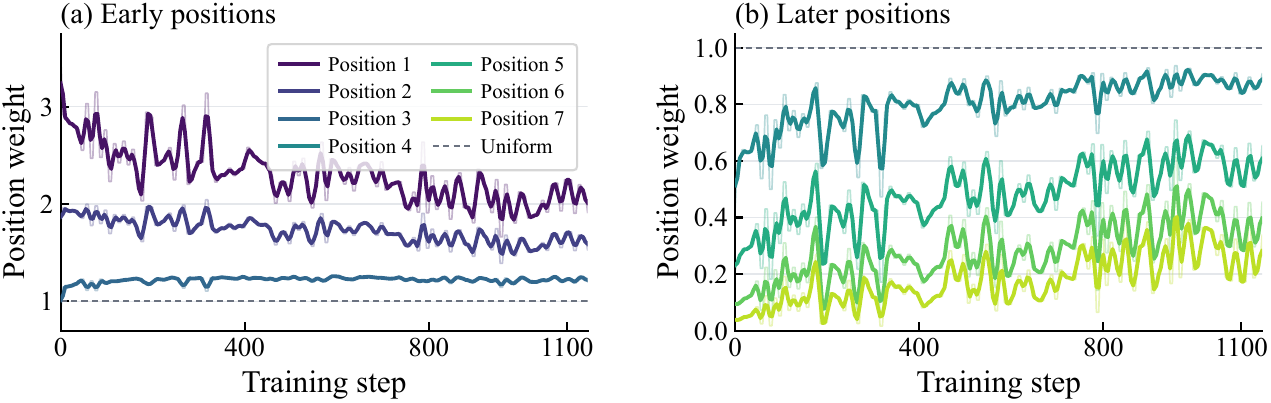}
    \caption{Evolution of the seven normalized proposal-position weights in our experimental setting. Early positions (1--3) and later positions (4--7) are separated for clarity. Faint curves show logged values, solid curves show a centered three-point moving average, and the dashed line denotes uniform weighting ($\bar w_i=1$). Emphasis initially favors early positions and gradually shifts toward later positions as the acceptance profile evolves. The panels use different vertical scales.}
    \label{fig:position_weight_evolution}
\end{figure}

Figure~\ref{fig:position_weight_evolution} shows how the normalized
position weights evolve during training. Early proposal positions
receive larger weights initially, reflecting their stronger influence
on prefix survival. As their acceptance improves, the weighting
gradually shifts toward later positions, which become the remaining
bottlenecks to verified progress. This behavior illustrates how the
proposed objective adapts its optimization focus as the acceptance
profile evolves.

\subsection{Candidate-Refinement Head Details}
\label{app:head_implementation}

We augment the frozen BRISK backbone with a lightweight recurrent head that
refines the base proposal scores at the $M=N-1=7$ speculative positions. At
position $j$, the head operates on the ordered candidate set
\begin{equation}
    \mathcal{S}_j = \operatorname{TopK}(q_j),
    \qquad K=16,
\end{equation}
where $q_j$ is the base proposal distribution. The head only reranks these
candidates and never expands the candidate set.

The recurrent state is initialized from the anchor hidden state and anchor
token embedding. At each proposal position, its input concatenates a
64-dimensional projection of the MASK hidden state, a 64-dimensional
projection of the previous-token embedding, a 16-dimensional position
embedding, and a 16-dimensional projection of five base-proposal confidence
features. These features contain the Top-1--Top-2 and Top-1--Top-4 logit gaps,
indicators of whether the previous token is the Top-1 token or belongs to the
Top-4, and the Top-1--previous-token logit gap. The resulting 160-dimensional
input is processed by a bias-free minimal GRU with a 128-dimensional state.

For each candidate $v_{j,k}\in\mathcal{S}_j$, the head predicts a gated,
rank-64 residual over the base logit,
\begin{align}
    d_{j,k}
      &= \left\langle W_s s_j,\,W_c E_{\mathrm{out}}(v_{j,k})\right\rangle,
      \\
    \gamma_j
      &= \sigma(\alpha)\,
         \sigma\!\left(w_g^\top[s_j;\phi_j]+b_g\right), \\
    \widetilde{\ell}_{j,k}
      &= \ell^q_{j,k}+\gamma_j d_{j,k},
    \label{eq:head_residual_logits}
\end{align}
where $E_{\mathrm{out}}$ denotes the frozen LM-head embedding and $\phi_j$
denotes the confidence features. The input embedding, LM head, and BRISK
backbone remain frozen. The refinement head contains 1.89M trainable
parameters.

We train the head on the correctness-filtered self-generated trajectories
described above. Previous-token feedback follows the actual base proposal
trajectory. The target at each reached position is the verifier distribution
restricted to and renormalized over the same Top-16 candidate set. If the
first rejected proposal occurs at position $b$, the loss covers the accepted
prefix and position $b$; for an all-pass event, it covers all seven proposal
positions. Positions after the first rejection are excluded.

Let $p_{\mathcal{S},j}$ be the verifier distribution, $q_{\mathcal{S},j}$ the
base distribution, and $r_j=\operatorname{softmax}(\widetilde{\ell}_j)$ the
refined distribution, all conditional on $\mathcal{S}_j$.
The distribution $r_j$ is used for training; verification uses the
actual proposal law induced by the configured inference selection rule.
The training objective is
\begin{equation}
\begin{aligned}
    \ell_{e,j}
    &= \operatorname{CE}(p_{\mathcal{S},e,j},r_{e,j})\quad + \ \beta D_{\mathrm{KL}}(q_{\mathcal{S},e,j}\,\|\,r_{e,j}),\\
    \mathcal{L}_{\mathrm{head}}
    &= \frac{\sum_{e,j}m_{e,j}\ell_{e,j}}{\sum_{e,j}m_{e,j}},
    \qquad \beta=0.01.
\end{aligned}
\label{eq:head_training_loss}
\end{equation}
Here $\operatorname{CE}$ is cross-entropy and $m_{e,j}$ is the
reached-position mask. We take one global mean over
valid positions, use uniform position weights, and apply no additional weight
to the first rejected position. Table~\ref{tab:head_implementation_details}
summarizes the implementation settings.

\begin{table}[ht]
\centering
\caption{Candidate-refinement head configuration.}
\label{tab:head_implementation_details}
\small
\setlength{\tabcolsep}{5pt}
\renewcommand{\arraystretch}{0.95}
\begin{tabular}{lc}
\toprule
Setting & Value \\
\midrule
Proposal positions / candidate size & $7$ / $16$ \\
Recurrent cell / state dimension & minimal GRU / $128$ \\
Hidden and token projection dimensions & $64$ / $64$ \\
Position / feature projection dimensions & $16$ / $16$ \\
Candidate residual rank & $64$ \\
Trainable parameters & $1.89$M \\
Training data & 326 prompts / 165,515 events \\
Supervision & accepted prefix + first rejection \\
Optimizer / learning rate & AdamW / $3{\times}10^{-4}$ \\
Anchor KL coefficient & $0.01$ \\
Trainable / forward precision & FP32 / BF16 \\
Training hardware & $1{\times}$A100 \\
\bottomrule
\end{tabular}
\end{table}
\subsection{Algorithm}
\begin{algorithm}[h]
\caption{Overview of BRISK-DLM training and inference.}
\label{alg:brisk}
\footnotesize

\begin{minipage}[t]{0.48\linewidth}
\textbf{(a) Training}
\begin{algorithmic}[1]
\Require Initial DLM $\theta_0$, prompt--answer pairs $\mathcal{D}$

\State $\mathcal{D}_{\mathrm{on}}\gets\emptyset$, $\theta\gets\theta_0$
\For{$(c,y)\in\mathcal{D}$}
    \State Generate responses $\mathcal{S}_c$ with $\theta_0$
    \State Add correct responses in $\mathcal{S}_c$
    to $\mathcal{D}_{\mathrm{on}}$
\EndFor

\While{not converged}
    \State Sample $\mathcal{B}\subset\mathcal{D}_{\mathrm{on}}$
    \State Compute proposal losses and accept profile $\mathbf{a}$
    \State Compute $\bar{\mathbf{w}}$ from $\mathbf{a}$ using Eq.~(3)
    \State Update $\theta$ with $\mathcal{L}_{\mathrm{BRISK}}$
\EndWhile

\Statex \textbf{Verifier-distilled corrector}
\State Freeze $\theta$ and collect policy-reached blocks
\State Obtain verifier targets on reached positions
\State Train corrector $\psi$ using Eq.~\eqref{eq:head_training_loss}
\State \Return $\theta,\psi$
\end{algorithmic}
\end{minipage}
\hfill
\begin{minipage}[t]{0.48\linewidth}
\textbf{(b) Inference}
\begin{algorithmic}[1]
\Require Trained DLM $\theta$, corrector $\psi$, prefix $c$

\State Compute $\{m_i,\ell_i^q,q_i\}_{i=1}^{M}$ and $h_c$
       \textbf{in parallel}
\State $x_0\gets$ token preceding the proposal block

\State $u_0 \gets \mathrm{Init}_{\psi}(h_c, E(x_0))$

\For{$i=1,\ldots,M$}
    \State $\mathcal{C}_i\gets\mathrm{TopK}(q_i,K)$
    \State $u_i\gets
    f_\psi(u_{i-1},h_c,m_i,E(x_{i-1}),\phi_i)$
    \State Refine logits on $\mathcal{C}_i$ using Eq.~(5)
    \State $x_i\gets\mathrm{Select}(\mathcal{C}_i,\tilde{\ell}_{\psi,i})$
\EndFor

\State Verify $x_{1:M}$ using $(p_i,\pi_i)$ as in
       \S\ref{sec:inference_head}
\State Commit the accepted prefix and correction;
       follow the ISD stopping rule
\State \Return updated prefix
\end{algorithmic}
\end{minipage}

\end{algorithm}

\subsection{Evaluation Details}
BRISK-DLM and I-DLM use identical serving configurations, with inference implemented in SGLang and block selection in Triton. Main-text comparisons use aggregate TPS; Appendix~\ref{app:h200_serving} additionally reports median request TPS and the complete hardware configuration. Figure annotations summarize paired throughput ratios using the geometric mean across concurrency.

\section{Detailed Serving Results}
\label{app:serving_results}

\subsection{Protocols and comparison scope}
Main serving results use aggregate generated-model-token throughput, equal to generated model tokens divided by elapsed wall-clock time. Each request generates exactly 2,048 model tokens and concurrency is $C\in\{1,2,4,8,16,32\}$. Geometric means average paired ratios, not raw TPS values. Quality is assessed separately using the task-specific metrics in Table~\ref{tab:quality}; a fixed-output throughput benchmark is not an accuracy or time-to-correct-answer measurement.

H200 comparisons use one GPU/TP1 for 8B I-DLM, BRISK, and Qwen3, and two GPUs/TP2 for the corresponding 32B models. LLaDA-2.0/2.1-flash uses four H200s/TP4; other listed H200 systems use one H200/TP1. The panel therefore compares native serving instances. DFlash OOD-thinking and native NT are separate operating modes. D-PACE uses the same target, runtime, and numeric path as the corresponding DFlash operating point, replacing only the drafter. The system optimization study in Appendix~\ref{app:head_system_optimization} uses a different client-visible-token counter and includes prefill exposures in its TPF denominator.

\subsection{Complete H200 Results}
\label{app:h200_serving}

Table~\ref{tab:h200_full} reports the aggregate TPS values used in the main figure. The 8B geometric-mean gains relative to I-DLM are 22.1\%, 22.9\%, and 23.1\% on GSM8K, MATH-500, and HumanEval; the 32B gains are 32.1\%, 35.9\%, and 21.8\%.

\begin{table}[t]
\centering
\caption{\textbf{Complete H200 serving results.} GPU/TP denotes the number of GPUs and tensor-parallel degree; each result cell reports aggregate TPS / median request TPS in generated model tokens/s. Rankings use the aggregate-TPS component: $^{*}$ denotes the best result in each column, underlining denotes the second-best result, and gray rows denote BRISK-DLM.}
\label{tab:h200_full}
\scriptsize
\setlength{\tabcolsep}{2.8pt}
\renewcommand{\arraystretch}{0.90}
\begin{adjustbox}{max width=\textwidth}
\begin{tabular}{lcrrrrrr}
\toprule
\multicolumn{8}{l}{\textbf{(a) GSM8K}} \\
Method & GPU/TP & \multicolumn{1}{l}{$C=1$} & \multicolumn{1}{l}{$C=2$} & \multicolumn{1}{l}{$C=4$} & \multicolumn{1}{l}{$C=8$} & \multicolumn{1}{l}{$C=16$} & \multicolumn{1}{l}{$C=32$} \\
\midrule
I-DLM-8B & 1/1 & 438.4/438.4 & 752.8/450.2 & 1539.0/403.5 & 2489.4/340.4 & 3858.4/293.9 & 4893.2/181.8 \\
\rowcolor{gray!15}\textbf{BRISK-DLM-8B (Ours)} & 1/1 & 566.1/566.1 & 1014.3/517.1 & 1909.0/464.9 & 2873.3/408.3 & 4447.6/337.1 & 5652.0/192.4 \\
DFlash OOD-thinking & 1/1 & 392.5/392.5 & 782.2/447.6 & 1303.1/371.5 & 2337.2/330.1 & 3835.1/297.1 & 4906.2/195.9 \\
DFlash native NT & 1/1 & 878.4/878.4$^{*}$ & \underline{1612.5/898.3} & \underline{2784.2/887.0} & \underline{4531.7/823.6} & \underline{6611.0/696.3} & \underline{10046.0/454.9} \\
D-PACE-DFlash OOD-thinking & 1/1 & 389.1/389.1 & 721.0/427.3 & 1441.5/384.0 & 2616.9/354.8 & 4026.7/323.0 & 5696.7/207.7 \\
D-PACE-DFlash native NT & 1/1 & 803.2/803.2 & 1976.2/1107.3$^{*}$ & 3467.8/1000.4$^{*}$ & 5604.2/838.8$^{*}$ & 8451.3/619.3$^{*}$ & 11074.1/441.5$^{*}$ \\
Qwen3-8B & 1/1 & 183.0/183.0 & 344.1/172.1 & 677.6/169.4 & 1282.8/160.4 & 2495.0/156.1 & 4391.1/137.4 \\
SDAR-8B & 1/1 & 302.6/302.6 & 313.7/240.6 & 304.3/101.2 & 698.0/125.2 & 642.9/62.1 & 2201.0/86.7 \\
\addlinespace[1pt]
LLaDA-2.1-mini (16B) & 1/1 & 476.9/476.9 & 501.0/388.8 & 213.5/60.2 & 453.9/67.1 & 950.2/68.0 & 1367.6/46.4 \\
SDAR-30B-A3B & 1/1 & 309.1/309.1 & 276.0/187.3 & 85.1/22.7 & 173.5/23.2 & 406.0/27.5 & 767.6/25.9 \\
LLaDA-2.0-flash (100B) & 4/4 & 686.4/686.4 & 863.1/778.1 & 919.1/280.9 & 684.7/103.6 & 577.2/40.9 & 1160.1/37.4 \\
LLaDA-2.1-flash (100B) & 4/4 & \underline{852.6/852.6} & 1429.2/714.7 & 1168.8/388.6 & 1494.5/225.9 & 1143.4/78.0 & 1264.8/40.7 \\
I-DLM-32B & 2/2 & 119.9/119.9 & 254.3/127.8 & 418.8/113.1 & 762.3/99.2 & 1135.1/75.4 & 1491.0/49.8 \\
\rowcolor{gray!15}\textbf{BRISK-DLM-32B (Ours)} & 2/2 & 167.1/167.1 & 302.1/158.5 & 561.5/158.2 & 997.3/146.5 & 1506.5/105.3 & 2058.5/72.3 \\
Qwen3-32B & 2/2 & 82.2/82.2 & 159.5/79.8 & 321.5/80.4 & 626.4/78.3 & 1210.9/75.7 & 2037.9/63.7 \\
\midrule
\multicolumn{8}{l}{\textbf{(b) MATH-500}} \\
Method & GPU/TP & \multicolumn{1}{l}{$C=1$} & \multicolumn{1}{l}{$C=2$} & \multicolumn{1}{l}{$C=4$} & \multicolumn{1}{l}{$C=8$} & \multicolumn{1}{l}{$C=16$} & \multicolumn{1}{l}{$C=32$} \\
\midrule
I-DLM-8B & 1/1 & 406.7/406.7 & 824.4/422.5 & 1467.7/408.2 & 2359.2/336.6 & 3375.9/247.5 & 4437.6/161.7 \\
\rowcolor{gray!15}\textbf{BRISK-DLM-8B (Ours)} & 1/1 & 516.9/516.9 & 998.4/481.6 & 1689.5/437.2 & 2950.0/399.3 & 4345.2/280.9 & 5367.3/182.9 \\
DFlash OOD-thinking & 1/1 & 377.8/377.8 & 706.3/372.8 & 1248.8/368.8 & 2090.4/313.2 & 3589.1/260.7 & 4454.7/167.8 \\
DFlash native NT & 1/1 & \underline{808.0/808.0} & 1563.7/973.0$^{*}$ & 2950.6/874.7$^{*}$ & 3857.8/685.5$^{*}$ & \underline{4992.5/504.8} & 7995.3/395.7$^{*}$ \\
D-PACE-DFlash OOD-thinking & 1/1 & 426.7/426.7 & 677.7/347.9 & 1303.4/384.9 & 2256.1/319.0 & 3550.7/251.4 & 4742.5/172.8 \\
D-PACE-DFlash native NT & 1/1 & 842.4/842.4$^{*}$ & \underline{1507.8/765.8} & \underline{2814.4/799.7} & \underline{3300.1/715.5} & 5487.4/510.4$^{*}$ & \underline{7865.5/399.6} \\
Qwen3-8B & 1/1 & 184.3/184.3 & 348.9/174.5 & 686.7/171.7 & 1321.1/165.2 & 2467.9/154.4 & 4327.8/135.4 \\
SDAR-8B & 1/1 & 304.7/304.7 & 308.9/232.9 & 383.8/137.3 & 668.7/114.2 & 1305.3/111.1 & 1902.3/78.5 \\
\addlinespace[1pt]
LLaDA-2.1-mini (16B) & 1/1 & 501.8/501.8 & 331.5/332.6 & 303.5/89.1 & 324.6/46.1 & 668.1/46.6 & 766.1/23.9 \\
SDAR-30B-A3B & 1/1 & 306.6/306.6 & 236.4/199.9 & 81.0/21.7 & 167.3/22.4 & 410.0/27.7 & 753.0/25.9 \\
LLaDA-2.0-flash (100B) & 4/4 & 475.1/475.1 & 272.8/136.4 & 543.9/154.1 & 461.3/66.1 & 484.4/32.7 & 883.9/28.5 \\
LLaDA-2.1-flash (100B) & 4/4 & 596.0/596.0 & 749.5/374.8 & 676.6/189.7 & 528.3/99.5 & 467.8/30.6 & 852.2/27.4 \\
I-DLM-32B & 2/2 & 124.6/124.6 & 239.6/123.4 & 438.8/118.5 & 764.8/101.1 & 1115.4/74.0 & 1448.3/48.0 \\
\rowcolor{gray!15}\textbf{BRISK-DLM-32B (Ours)} & 2/2 & 161.6/161.6 & 340.9/172.9 & 660.1/167.6 & 1021.7/140.1 & 1434.2/100.3 & 1909.7/71.7 \\
Qwen3-32B & 2/2 & 81.6/81.6 & 159.7/79.9 & 319.7/79.9 & 629.2/78.7 & 1207.1/75.5 & 1890.6/59.1 \\
\midrule
\multicolumn{8}{l}{\textbf{(c) HumanEval}} \\
Method & GPU/TP & \multicolumn{1}{l}{$C=1$} & \multicolumn{1}{l}{$C=2$} & \multicolumn{1}{l}{$C=4$} & \multicolumn{1}{l}{$C=8$} & \multicolumn{1}{l}{$C=16$} & \multicolumn{1}{l}{$C=32$} \\
\midrule
I-DLM-8B & 1/1 & 536.3/536.3 & 905.0/498.1 & 1711.9/481.2 & 2573.4/389.1 & 3657.6/285.6 & 4798.3/170.5 \\
\rowcolor{gray!15}\textbf{BRISK-DLM-8B (Ours)} & 1/1 & 686.4/686.4 & 1114.9/561.1 & 2033.6/518.8 & 3125.5/399.4$^{*}$ & 4689.6/309.3$^{*}$ & \underline{5720.1/181.0} \\
DFlash OOD-thinking & 1/1 & 676.2/676.2 & 780.0/408.3 & 1765.9/476.5 & 2176.0/447.7 & 3579.7/305.2 & 4521.6/189.9 \\
DFlash native NT & 1/1 & 215.4/215.4 & 415.5/819.1 & 858.1/1263.3 & 1544.6/1128.2 & 2573.8/851.5 & 3967.8/545.4 \\
D-PACE-DFlash OOD-thinking & 1/1 & 516.3/516.3 & 817.6/468.9 & 1406.2/498.5 & 2301.4/418.3 & 3801.0/318.6 & 4850.1/198.2 \\
D-PACE-DFlash native NT & 1/1 & 657.6/657.6 & 1304.4/678.5 & 1458.2/1362.8 & 2838.8/1228.8 & \underline{4373.9/532.0} & 6468.0/533.0$^{*}$ \\
Qwen3-8B & 1/1 & 182.7/182.7 & 344.3/172.2 & 676.0/169.0 & 1271.7/159.0 & 2475.2/154.8 & 4327.0/135.4 \\
SDAR-8B & 1/1 & 275.8/275.8 & 280.5/212.3 & 314.0/109.4 & 731.8/136.0 & 1006.8/120.7 & 1851.7/57.9 \\
\addlinespace[1pt]
LLaDA-2.1-mini (16B) & 1/1 & 403.9/403.9 & 449.4/363.7 & 204.0/58.3 & 377.0/53.6 & 720.7/50.5 & 1076.2/37.1 \\
SDAR-30B-A3B & 1/1 & 254.8/254.8 & 244.2/177.8 & 89.7/24.6 & 171.0/23.3 & 398.0/27.6 & 761.1/26.6 \\
LLaDA-2.0-flash (100B) & 4/4 & 1581.5/1581.5 & \underline{1586.9/1165.1} & \underline{2162.4/741.0} & 2683.6/402.4 & 2054.1/139.3 & 3999.0/135.2 \\
LLaDA-2.1-flash (100B) & 4/4 & 1882.5/1882.5$^{*}$ & 1754.6/1346.0$^{*}$ & 2380.7/859.4$^{*}$ & \underline{3055.6/636.9} & 2343.0/159.0 & 4466.8/152.0 \\
I-DLM-32B & 2/2 & 121.2/121.2 & 228.2/120.4 & 443.2/118.1 & 744.5/108.9 & 1068.0/72.0 & 1412.7/47.6 \\
\rowcolor{gray!15}\textbf{BRISK-DLM-32B (Ours)} & 2/2 & 163.9/163.9 & 290.1/157.4 & 511.5/183.8 & 872.4/131.8 & 1273.0/87.3 & 1660.8/55.8 \\
Qwen3-32B & 2/2 & 81.5/81.5 & 159.2/79.6 & 319.9/80.0 & 627.5/78.4 & 1204.7/75.3 & 1644.2/51.4 \\
\bottomrule
\end{tabular}
\end{adjustbox}
\vspace{2pt}
\parbox{\textwidth}{\footnotesize Rows are grouped with the 8B systems first, followed by larger-scale references.}
\end{table}

\subsection{Tokens per Forward}
\label{app:tpf_results}

Table~\ref{tab:h200_tpf_c1} reports C=1 TPF over the same fixed 2,048-token measurement window. For DFlash and D-PACE-DFlash, TPF counts target-verification cycles and excludes draft-model forward cost, so it is a mechanism diagnostic rather than a complete measure of end-to-end cost. Qwen3 has TPF 1 by autoregressive definition.

\begin{table}[H]
\centering
\caption{H200 C=1 tokens per forward. A dash denotes an unavailable matched measurement.}
\label{tab:h200_tpf_c1}
\small
\setlength{\tabcolsep}{7pt}
\begin{tabular}{lrrr}
\toprule
Method & MATH-500 & GSM8K & HumanEval \\
\midrule
LLaDA-2.0-flash (100B) & 3.814 & 5.404 & 11.636 \\
LLaDA-2.1-flash (100B) & 4.697 & 6.628 & 13.838 \\
SDAR-8B & 1.824 & 1.770 & 1.618 \\
SDAR-30B-A3B & 1.680 & 1.817 & 1.411 \\
DFlash OOD-thinking & 3.089 & 3.141 & 5.404 \\
DFlash native NT & 6.502 & 7.087 & 1.724 \\
D-PACE-DFlash OOD-thinking & 3.448 & 3.070 & 4.096 \\
D-PACE-DFlash native NT & 6.759 & 6.502 & 9.421 \\
I-DLM-8B & 2.794 & 2.897 & 3.710 \\
\textbf{BRISK-DLM-8B (Ours)} & \textbf{3.497} & \textbf{4.264} & \textbf{4.534} \\
Qwen3-8B & 1.000 & 1.000 & 1.000 \\
I-DLM-32B & 2.048 & 1.967 & 1.990 \\
BRISK-DLM-32B & 2.734 & 2.805 & 2.713 \\
Qwen3-32B & 1.000 & 1.000 & 1.000 \\
\bottomrule
\end{tabular}
\end{table}

\begin{table}[H]
\centering
\caption{A100 MATH-500 throughput (aggregate TPS / median request TPS).
GPU/TP denotes GPU count and tensor-parallel degree. $^\dagger$ denotes TP2 at $C\leq8$ and TP4 at $C\geq16$.}
\label{tab:a100_serving}
\scriptsize
\setlength{\tabcolsep}{3pt}
\renewcommand{\arraystretch}{1.10}
\begin{adjustbox}{max width=\textwidth}
\begin{tabular}{lcrrrrrr}
\toprule
Method & GPU/TP & $C=1$ & $C=2$ & $C=4$ & $C=8$ & $C=16$ & $C=32$ \\
\midrule
SDAR-8B & 1/1 & 128.94 / 128.94 & 130.25 / 96.86 & 160.89 / 57.78 & 309.67 / 64.86 & 521.80 / 32.61 & 950.81 / 39.90 \\
SDAR-30B-A3B & 2/2 & 125.12 / 125.13 & 128.83 / 98.25 & 61.66 / 17.57 & 97.06 / 13.42 & 194.36 / 13.16 & 354.39 / 12.10 \\
DFlash OOD-thinking & 1/1 & 160.50 / 160.50 & 319.18 / 160.18 & 592.32 / 168.67 & 976.71 / 142.05 & 1466.81 / 106.81 & 1773.39 / 65.67 \\
DFlash native NT & 1/1 & 573.60 / 573.60 & 1008.56 / 531.07 & 1142.62 / 385.19 & 1413.40 / 328.30 & 2411.27 / 254.63 & 2937.99 / 146.33 \\
I-DLM-8B & 1/1 & 182.19 / 182.19 & 333.99 / 169.59 & 579.19 / 160.42 & 965.88 / 137.14 & 1491.94 / 103.58 & 1723.26 / 59.89 \\
\textbf{BRISK-DLM-8B (Ours)} & 1/1 & 231.56 / 231.56 & 385.00 / 195.49 & 666.74 / 184.67 & 1207.78 / 171.49 & 1920.29 / 133.32 & 2084.28 / 72.44 \\
Qwen3-8B & 1/1 & 73.48 / 73.48 & 143.76 / 71.88 & 278.86 / 69.72 & 511.22 / 63.91 & 914.98 / 57.20 & 1579.37 / 49.37 \\
I-DLM-32B & 2/2; 4/4$^\dagger$ & 57.19 / 57.19 & 101.72 / 50.99 & 181.62 / 47.93 & 303.05 / 40.61 & 751.97 / 49.47 & 956.14 / 31.95 \\
\textbf{BRISK-DLM-32B (Ours)} & 2/2; 4/4$^\dagger$ & 75.61 / 75.61 & 137.15 / 68.66 & 256.24 / 67.65 & 389.63 / 56.99 & 970.71 / 68.52 & 1247.33 / 46.08 \\
Qwen3-32B & 2/2; 4/4$^\dagger$ & 34.46 / 34.46 & 68.23 / 34.12 & 134.97 / 33.75 & 264.11 / 33.02 & 794.01 / 49.63 & 1454.08 / 45.46 \\
\bottomrule
\end{tabular}
\end{adjustbox}
\end{table}

BRISK-DLM-8B exceeds same-scale I-DLM at every listed concurrency, with a 21.9\% geometric-mean gain. For 32B, the means are 34.1\% in the TP2 regime and 29.8\% in the TP4 regime, each relative to I-DLM under the same allocation. These results do not imply superiority to every other method: at $C=32$/TP4, Qwen3-32B achieves 1454.08 TPS versus 1247.33 for BRISK-DLM-32B. This A100 panel is separate from the 64-prompt head execution study.

\subsection{Stride-size sensitivity}
Table~\ref{tab:stride_sensitivity} reports the complete throughput
sweep over $N\in\{4,8,16\}$ under the same H200 serving setup.

\begin{table}[h]
\centering
\caption{\textbf{Stride-size sensitivity on MATH-500.}
Total serving throughput (model tokens/s) for Original I-DLM and
BRISK across proposal widths and concurrency levels.
GM gain denotes the geometric mean of the BRISK/I-DLM throughput
ratio over all six concurrency levels.}
\label{tab:stride_sensitivity}
\scriptsize
\setlength{\tabcolsep}{3.0pt}
\renewcommand{\arraystretch}{1.05}
\begin{adjustbox}{max width=\columnwidth}
\begin{tabular}{lrrrrrrr}
\toprule
Method
& $C=1$ & $C=2$ & $C=4$ & $C=8$ & $C=16$ & $C=32$
& GM gain \\
\midrule
I-DLM $N=4$
& 427.6 & 772.8 & 1484.1 & 2430.8 & 4157.8 & 6239.1 & -- \\
BRISK $N=4$
& 457.1 & 904.2 & 1705.0 & 2982.9 & 4736.8 & 7057.7 & +14.7\% \\
\addlinespace[1pt]

I-DLM $N=8$
& 406.7 & 824.4 & 1467.7 & 2359.2 & 3375.9 & 4437.6 & -- \\
BRISK $N=8$
& 516.9 & 998.4 & 1689.5 & 2950.0 & 4345.2 & 5367.3 & \textbf{+22.9\%} \\
\addlinespace[1pt]

I-DLM $N=16$
& 398.7 & 737.8 & 1290.2 & 1891.4 & 2247.3 & 2619.9 & -- \\
BRISK $N=16$
& 482.9 & 828.5 & 1477.6 & 2306.3 & 2791.6 & 3347.7 & +20.2\% \\
\bottomrule
\end{tabular}
\end{adjustbox}
\end{table}

\subsection{Impact of Hardware} 
\label{sec:hardware_ablation}
We compare MATH-500 aggregate throughput on H200 and A100
(Table~\ref{tab:h200_full} and Table~\ref{tab:a100_serving}).
With one GPU, BRISK-DLM-8B outperforms I-DLM at all six concurrency
levels, with geometric-mean gains of 22.9\% on H200 and 21.9\% on A100.
For 32B, restricting the comparison to the common TP2 regime
($C\in\{1,2,4,8\}$) yields geometric-mean gains of 38.8\% on H200 and
34.1\% on A100.

\section{Top-\texorpdfstring{$K$}{K} Sensitivity Analysis}
\label{app:topk_analysis}

\paragraph{Evaluation setting.}
We evaluate inference support $K\in\{4,8,16,32\}$ with a frozen
8B BRISK backbone.
Fixed-state candidate statistics and natural-EOS generation are
evaluated on the full MATH-500 benchmark. Serving is evaluated on an
A100 at $C\in\{1,8,16\}$ with a fixed-output request bank.

\begin{table}[t]
\centering
\caption{\textbf{Top-$K$ serving throughput on A100 (TPS).}}
\label{tab:topk_details}
\begingroup

\setlength{\tabcolsep}{3pt}
\renewcommand{\arraystretch}{1.0}
\begin{tabular}{@{}rrrr@{}}
\toprule
$K$ & $C=1$ & $C=8$ & $C=16$\\
\midrule
4  & 91.70 & 706.19 & 1085.16\\
8  & 92.90 & 682.29 & 1076.39\\
16 & \textbf{94.22} & \textbf{741.38} & \textbf{1100.17}\\
32 & 92.47 & 707.18 & 1012.90\\
\bottomrule
\end{tabular}
\endgroup
\end{table}

\paragraph{Marginal utility beyond $K=16$.}
Expanding the support from $K=16$ to $K=32$ changes the selected-token
$\log p$ only from $-11.671$ to $-11.669$, while ranks 17--32 are
selected in 0.071\% of first-breaker states. We therefore report this
as a direct $K=16\!\rightarrow\!32$ comparison rather than as a sweep.
These results suggest that $K=16$ already contains nearly all
candidates that affect the corrector's decision: the additional ranks
increase candidate availability but rarely change the selected token.
Consequently, expanding to $K=32$ offers little room for further gains
with the fixed corrector used in this evaluation.

\paragraph{Serving throughput.}
Across $C=1$, $C=8$, and $C=16$, $K=16$ attains the highest
throughput, reaching 94.22, 741.38, and 1100.17 TPS, respectively.
It is 1.4--5.0\% faster than the best smaller-support setting at each
concurrency and 1.9--8.6\% faster than $K=32$. Thus, $K=16$ gives the
highest measured throughput across all evaluated loads.
\section{Inference-Head System Optimization}
\label{app:head_system_optimization}

\paragraph{Setup.}
We evaluate 64 MATH-500 prompts on one NVIDIA A100-SXM4-40GB (TP=1),
with exactly 2,048 client-visible tokens per request, stride $N=8$, and
$C\in\{1,8,32\}$. All settings share the trained backbone; both head-enabled settings use the same causal-head checkpoint. We report medians over six counterbalanced fresh-server runs per setting, computing request p95 within each run first. 
TPF divides client-visible tokens by request-level backbone
forward exposures, including prefill and actual active batch sizes in decoding.

\paragraph{From proposal efficiency to throughput.}
Both head implementations improve accepted-prefix length and TPF
(Table~\ref{tab:head_system_ablation}), but reference execution overhead outweighs this benefit and reduces TPS. Optimized execution increases TPS by 8.2\%, 9.5\%, and 9.1\% over head off at $C=1,8,32$. The optimized head retains similar
proposal efficiency while exceeding head-off TPS and lowering request p95 at
all three concurrency levels.

\begin{figure}[H]
    \centering
    \includegraphics[width=0.94\linewidth]{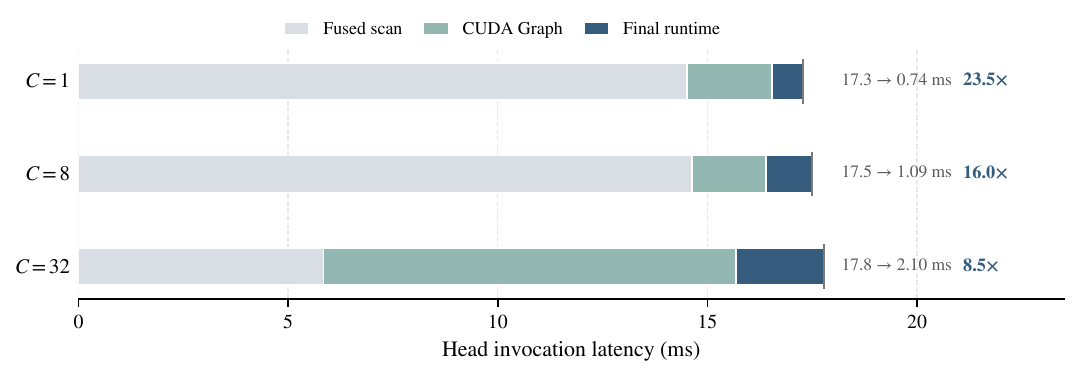}
    \caption{\textbf{Cumulative reduction in head execution cost.}
    Each bar reconstructs reference latency: two differences between successive measured paths (reference to fused scan, then to CUDA Graph replay), followed by the remaining runtime. These segments are not independent kernel timings. Labels show reference-to-optimized latency and speedup.}
    \label{fig:head_runtime_cumulative}
\end{figure}

\paragraph{Head execution cost.}
On the same recorded inputs, fused recurrent execution followed by CUDA Graph
replay reduces head-invocation latency by $23.5\times$, $16.0\times$, and
$8.5\times$ at $C=1,8,32$, respectively (Figure~\ref{fig:head_runtime_cumulative}).
These are head-runtime speedups; Table~\ref{tab:head_system_ablation} measures
end-to-end serving performance.

\begin{table}[H]
    \centering
    \small
    \caption{\textbf{End-to-end inference-head system ablation.}
    Absolute medians over six fresh-server runs per setting on 64 MATH-500
    prompts; request p95 is computed per run. Both head implementations use
    the same checkpoint.}
    \label{tab:head_system_ablation}
    \begingroup
    \setlength{\tabcolsep}{4pt}
    \renewcommand{\arraystretch}{1.10}
    \begin{tabular}{@{}clrrrr@{}}
        \toprule
        $C$ & Setting & \shortstack{Mean prefix\\$\uparrow$}
        & \shortstack{TPF\\$\uparrow$}
        & \shortstack{Total TPS\\$\uparrow$}
        & \shortstack{Request p95 (s)\\$\downarrow$} \\
        \midrule
        1 & Head Off & 4.20 & 3.37 & 90.0 & 30.20 \\
         & Causal Head (Reference) & 4.58 & 3.74 & 60.4 & 40.35 \\
         & \textbf{Causal Head (Optimized)} & 4.62 & 3.75 & \textbf{97.4} & \textbf{25.00} \\
        \midrule
        8 & Head Off & 4.23 & 3.39 & 682.1 & 24.72 \\
         & Causal Head (Reference) & 4.58 & 3.71 & 436.3 & 38.18 \\
         & \textbf{Causal Head (Optimized)} & 4.60 & 3.73 & \textbf{747.0} & \textbf{21.91} \\
        \midrule
        32 & Head Off & 4.26 & 3.43 & 1237.3 & 49.67 \\
         & Causal Head (Reference) & 4.57 & 3.71 & 1007.7 & 64.00 \\
         & \textbf{Causal Head (Optimized)} & 4.57 & 3.69 & \textbf{1350.3} & \textbf{47.42} \\
        \bottomrule
    \end{tabular}
    \endgroup
\end{table}

\end{document}